\documentclass[10pt,twocolumn]{article}

\usepackage{amssymb}
\usepackage{amsthm}
\usepackage{amsmath}
\usepackage{amsfonts}
\usepackage{float}
\usepackage{caption}
\usepackage{subcaption}
\usepackage{natbib}
\usepackage{tabularray}
\usepackage{booktabs}
\usepackage[dvipsnames]{xcolor}
\usepackage{mathtools}
\usepackage{lineno}
\usepackage[ruled,vlined,linesnumbered]{algorithm2e}

\SetCommentSty{mycommfont}

\begin{document}

\title{3C-FBI: A Combinatorial method using Convolutions for Circle Fitting in Blurry Images}

\author{Esteban Rom\'an Catafau$^1$ and Torbj\"orn E.M. Nordling$^{1,*}$\\
$^1$Mechanical Engineering, National Cheng Kung University\\
No. 1, University Road, Tainan City 701, Taiwan (R.O.C.)\\
$^*$Corresponding author: torbjorn.nordling@nordlinglab.org}

\date{}

\maketitle
            
\begin{abstract}

This paper addresses the fundamental computer vision challenge of robust circle detection and fitting in degraded imaging conditions. We present Combinatorial Convolution-based Circle Fitting for Blurry Images (3C-FBI), an algorithm that bridges the gap between circle detection and precise parametric fitting by combining (1) efficient combinatorial edge pixel (edgel) sampling and (2) convolution-based density estimation in parameter space.

We evaluate 3C-FBI across three experimental frameworks: (1) real-world medical data from Parkinson’s disease assessments (144 frames from 36 videos), (2) controlled synthetic data following established circle-fitting benchmarks, and (3) systematic analysis across varying spatial resolutions and outlier contamination levels. Results show that 3C-FBI achieves state-of-the-art accuracy (Jaccard index 0.896) while maintaining real-time performance (40.3 fps), significantly outperforming classical methods like RCD (6.8 fps) on a standard CPU (i7-10875H). It maintains near-perfect accuracy (Jaccard $\sim$ 1.0) at high resolutions (480×480) and reliable performance (Jaccard $>$ 0.95) down to 160×160 with up to 20\% outliers.

In extensive synthetic testing, 3C-FBI achieves a mean Jaccard Index of 0.989 across contamination levels, comparable to modern methods like Qi et al. (2024, 0.991), and surpassing RHT (0.964). This combination of accuracy, speed, and robustness makes 3C-FBI ideal for medical imaging, robotics, and industrial inspection under challenging conditions.


\end{abstract} 

\setcounter{secnumdepth}{0}

\section{Introduction}
\label{sec:intro}

Circle detection and fitting is a fundamental task in computer vision, with wide-ranging applications across various domains. 
This technique plays a crucial role in biometric authentication systems, particularly in iris segmentation \cite{ma2002iris, van2008novel, arvacheh2006iris}, and is instrumental in robotics for visual localization using circular markers \cite{xavier2005fast, glover2016event, teixido2012two}.
Moreover, detecting circles aids in recognizing man-made objects like traffic signs, a critical component for enabling autonomous navigation \cite{koresh2019computer, satti2020enhancing}. 
Its utility extends to industrial inspection, remote sensing imagery analysis \cite{ok2015circular, cai2014automatic}, and medical applications, including the monitoring of conditions like Parkinson's Disease (PD).

In computer vision, a clear distinction must be made between circle detection and circle fitting. 
Both tasks typically operate within a three-dimensional parameter space defined by the circle’s center coordinates (x, y) and radius r. 
Circle detection seeks to identify the presence and approximate location of circular patterns within an image, commonly employing methods such as the Hough Transform, randomized sampling techniques, or, more recently, convolutional neural networks. 
In contrast, circle fitting focuses on computing the most accurate circular representation for a given set of points, usually through mathematical optimization techniques like least squares fitting, to achieve sub-pixel precision. 
The fitting process minimizes the geometric distance between the data points and the estimated circle, often formalized as:
\begin{equation}
\min_{(x_c, y_c, r)} \sum_{i=1}^{n} \left(\sqrt{(x_i - x_c)^2 + (y_i - y_c)^2} - r\right)^2
\end{equation}
where $(x_c, y_c)$ is the circle center, $r$ is the radius, and $(x_i, y_i)$ are the input points. 

Despite numerous advancements in the field, accurate circle detection remains challenging, particularly in scenarios involving image distortions, variable lighting conditions, and complex backgrounds. 
These challenges are compounded by three main factors: (1) the sensitivity of edge detection to noise and blur, (2) the computational complexity of parameter space exploration, and (3) the presence of outliers and incomplete circle contours in real-world scenarios. 
This paper addresses these challenges by introducing 3C-FBI, a combinatorial algorithm for accurate circle detection in challenging scenarios. 

\subsection{Key Contributions}
Our study presents the following contributions:
\begin{itemize}
    \item Introduction of 3C-FBI, a new algorithm leveraging edgel analysis to enhance precision in identifying circular patterns. 
    The algorithm combines combinatorial sampling with convolution-based density estimation in parameter space, achieving robust detection even in degraded imaging conditions.
    
    \item Demonstration of improved circle detection accuracy in real-world scenarios with image distortions through an adaptive thresholding mechanism that automatically adjusts to varying image quality conditions.
    
    \item Comparative analysis of 3C-FBI against traditional and recent methods, showcasing its enhanced precision and efficiency. 
    Our method maintains real-time processing capability (40.31 fps) while achieving accuracy comparable to state-of-the-art approaches.
\end{itemize}

\subsection{Impact of Contributions}
The 3C-FBI algorithm significantly improves circle detection accuracy in challenging real-world scenarios, offering enhanced precision and efficiency. 
Our method demonstrates particular robustness to three common challenges: motion blur, variable lighting conditions, and partial occlusions. 
The algorithm maintains stable performance with up to 20\% outlier contamination and significant spatial discretization effects, making it especially suitable for real-time applications in robotics, medical imaging, and industrial quality control.

\subsection{Paper Organization}
The remainder of this paper is organized as follows: 
Section \textbf{Related Work} reviews related work in circle detection. 
Section \textbf{Materials and Methods} describes the dataset and the 3C-FBI algorithm. 
Section \textbf{Results} presents our experimental findings. 
Finally, Section \textbf{Conclusion} summarizes the study's contributions and outlines future research directions.

\section{Related Work}
\label{sec:RelWork}

Circle detection and fitting techniques have evolved significantly in computer vision, progressing from classical geometric approaches to modern machine learning methods. 
Circle detection and fitting methods has recent been reviewed in NNNN 
We organize the remaining section using their classification
this review by the fundamental approaches taken to solve this challenge.

\subsection{Geometric Methods}
The Circle Hough Transform (CHT) \cite{duda1972use} laid the foundation for reliable circle detection by mapping edgels into a parameter space. 
It identifies circular shapes by accumulating votes in a three-dimensional space defined by center coordinates $(x_c, y_c)$ and radius $r$, based on the equation $(x - x_c)^2 + (y - y_c)^2 = r^2$. 
Although robust, CHT is computationally intensive, prompting the development of more efficient alternatives. 
One such improvement is the Random Hough Transform (RHT) \cite{xu1990new}, which applies statistical sampling to lower processing costs without compromising accuracy. 
Building on this, Chen and Chung \cite{chen2001efficient} proposed the Random Circle Detection (RCD) algorithm, which streamlines the process by validating circles using randomly chosen triplets of edgels.

\subsection{Optimization-Based Approaches}

Atherton and Kerbyson \cite{atherton1999size} made a notable contribution by introducing convolution-based operators tailored for circle detection, enhancing both the precision and efficiency of the process. 
Ladrón de Guevara et al. \cite{ladron2011robust} proposed the Robust Fitting of Circle Arcs (RFCA) method, which employs an iterative strategy to minimize geometric errors while accounting for the presence of outliers.

More recent approaches in optimization have emphasized resilience to noise and outliers. 
For example, Nurunnabi et al. \cite{nurunnabi2018robust} merged robust Principal Component Analysis (PCA) with algebraic fitting, achieving dependable performance under up to 44\% outlier contamination. 
Expanding on this idea, Guo et al. \cite{guo2019iterative} combined Taubin’s algebraic fitting with a Median Absolute Deviation (MAD) based filtering technique to enhance robustness in noisy environments. 
Greco et al. \cite{greco2023impartial} further improved outlier handling by introducing an impartial trimming strategy supplemented with a reweighting phase.

The most recent development comes from Qi et al. \cite{qi2024robust}, who presented a robust weighted Hyper method that integrates M-estimators with the classical Hyper approach and incorporates the Black-Rangarajan duality framework. 
Their method adaptively suppresses outliers by assigning them negligible or zero weights. 
Their results showed exceptional accuracy and consistency, particularly in root mean squared error and repeatability, making the method well-suited for demanding tasks such as precise alignment in particle accelerators.

These innovations are grounded in foundational techniques, including the original Hyper method \cite{al2009error}, the Least Trimmed Symmetry Distance \cite{wang2003using}, and applications to laser scanning contexts \cite{nurunnabi2018robust, cai2013combined}.

\subsection{Evolutionary and Hybrid Approaches}
Alternative optimization methods have also been explored using evolutionary algorithms.
Ayala et al. \cite{ayala2006circle} showcased the effectiveness of genetic algorithms in achieving sub-pixel precision for circle detection tasks.
Jia et al. \cite{jia2011fast} and Chung et al. \cite{chung2012efficient} introduced gradient-based refinements that improved localization by leveraging edgel projection and applying constrained sampling techniques.
Promising results have also been observed in hybrid strategies that blend multiple methodologies—for example, the work by Lestriandoko et al. \cite{Lestriandoko2017Circle}, which combined edge detection filtering with the Circle Hough Transform (CHT).

\subsection{Deep Learning Methods}
In recent years, there has been a growing trend toward utilizing deep learning techniques for circle detection.
Ferede et al. \cite{ferede2019channel} introduced a detector enhanced by channel features within an SSD framework, while Kamble et al. \cite{kamble2019convolutional} showcased the capabilities of convolutional neural networks (CNNs) in tracking circles under dynamic conditions.
These approaches are particularly well-suited for complex, real-world environments, although they generally demand large volumes of labeled training data.

Each category of method comes with its own strengths and drawbacks.
Geometric methods are known for their robustness but often involve high computational costs.
Optimization-based methods offer accurate fitting but can be sensitive to initialization and outliers.
Deep learning solutions handle visual complexity effectively but rely heavily on extensive datasets.
The 3C-FBI algorithm proposed in this work draws on the strengths of these established methods while overcoming their key limitations through a combinatorial design.
The algorithm presented herein, 3C-FBI, is based on the combinatorial idea previously outlined in our preliminary work \cite{romancibica}, but the estimation of the circle using least squares is replaced by a convolution-based density estimation and adaptive thresholding for enhanced performance.

 \section{Materials and Methods}
\label{sec:Methods}

The circle detection algorithm developed for this research was focused on the toe-tapping assessment from the Movement Disorder Society-Unified Parkinson's Disease Rating Scale (MDS-UPDRS) \cite{goetz2008movement}, which serves as a standardized instrument for evaluating motor and non-motor manifestations in Parkinson's disease patients.
The assessment protocol involves having the patient maintain an upright seated position with both feet flat on the ground while performing toe-tapping movements on each foot individually.
Each foot must perform 10 tapping motions with maximum amplitude and speed, during which clinicians evaluate parameters including velocity, range of motion, interruptions, cessations, and progressive amplitude reduction.
\subsection{Experimental Setup}
\begin{figure}[tb]
\centering
\includegraphics[width=0.4\textwidth]{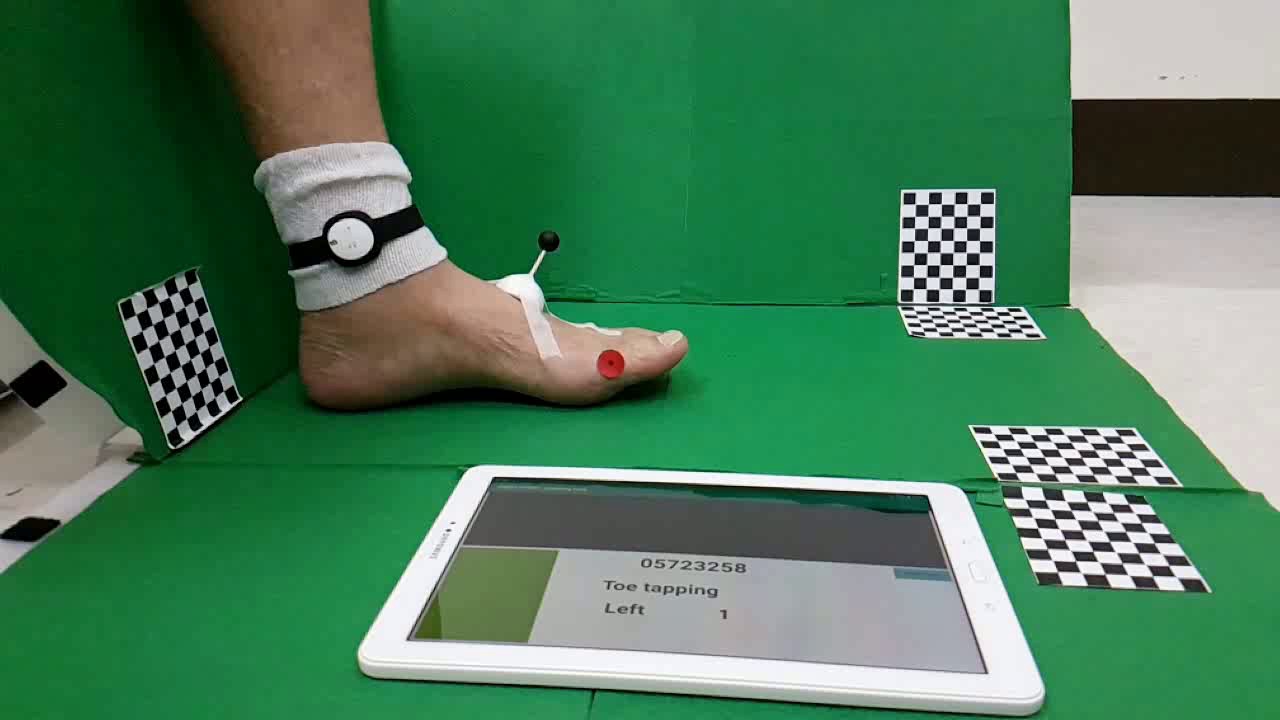}
\caption{Side-view capture of the foot using a laterally positioned camera. A black spherical marker was secured to the foot's upper surface to serve as a dimensional reference. Multiple chessboard calibration targets were strategically positioned within the imaging field to support computer vision processing. A red marker containing a black dot was placed on the foot's lateral aspect for trajectory tracking purposes. Furthermore, an inertial sensor was mounted on the ankle to facilitate multi-sensor data fusion.}
\label{fig:sidefoot}
\end{figure}

For the toe-tapping experiments, foot motion was captured using dual Samsung S7 cameras (12 MP, f/1.7, 26 mm), each featuring a 1/2.6 inch sensor with 1.4 $\mu$m pixel dimensions.
The cameras were strategically positioned with one facing the foot frontally and another placed laterally.
This study's analysis exclusively utilized footage from the side-mounted camera.
The detailed experimental methodology is described in \citet{ashyani2022digitization}.

\subsection{Dataset}
\begin{figure*}[!htb]
\centering
\includegraphics[width=\textwidth]{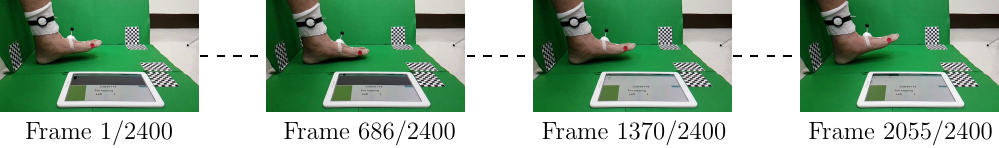}
\caption{Representative frames sampled from a standard video sequence for analytical processing.}
\label{fig:frameextraction}
\end{figure*}

The experimental dataset consisted of video recordings from 18 participants, comprising 12 individuals diagnosed with PD and 6 neurologically healthy control subjects.
Side-view recordings were captured for each participant's feet individually, generating a total of 36 video sequences.
From each video, four representative frames were selected for analysis, producing an overall dataset of 144 frames.
Video acquisition was performed at 720p resolution with a 240 frames per second capture rate, where each recording spanned 10 seconds (totaling 2400 frames).
The analytical frames were systematically sampled at regular intervals throughout each sequence, as demonstrated in Figure \ref{fig:frameextraction}.

\subsection{Black sphere cropping}
To optimize the circle detection process for the black spherical marker in each frame, the sphere was segmented from the surrounding environment to minimize computational demands.
A representative background region was extracted to characterize its chromatic properties, involving computation of minimum, maximum, mean, and standard deviation statistics across all RGB channels.
These statistical parameters enabled implementation of a color-based thresholding procedure for initial sphere isolation.
\begin{figure}[!htb]
\centering
\begin{subfigure}[b]{0.23\textwidth}
\centering
\includegraphics[width=\textwidth]{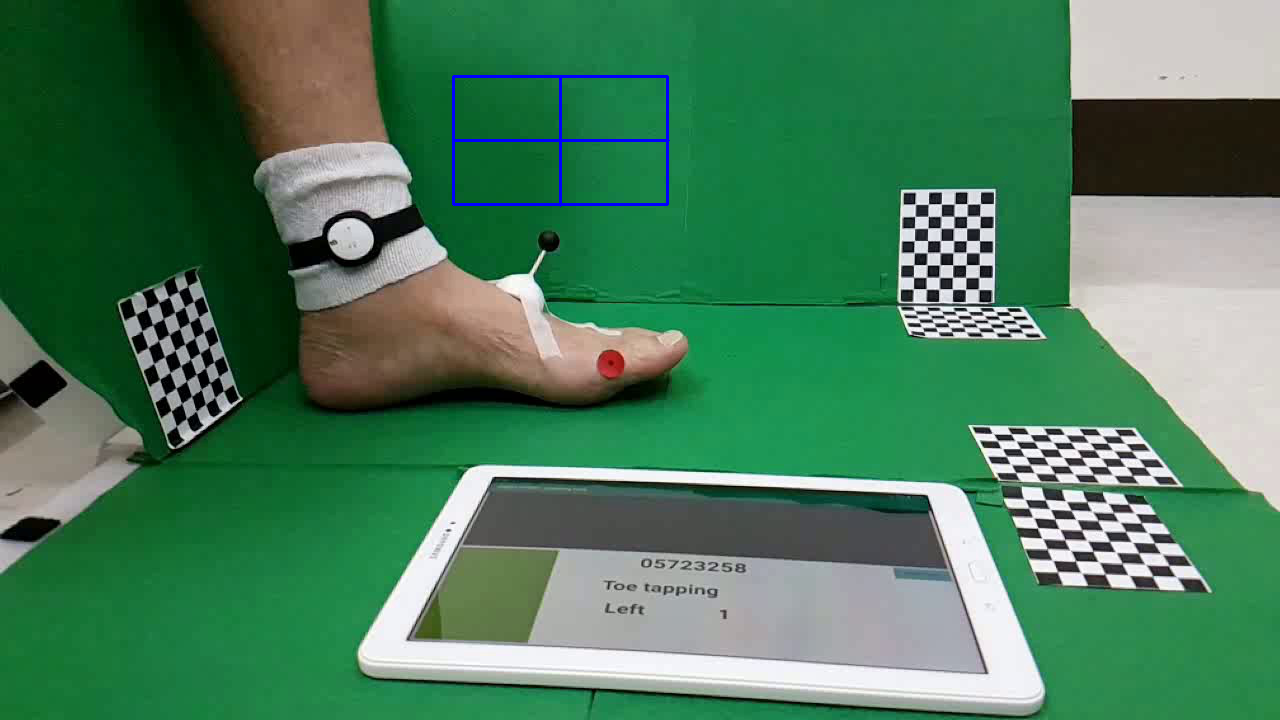}
\caption{Background region extraction}
\label{fig:GreenCrop}
\end{subfigure}
\begin{subfigure}[b]{0.23\textwidth}
\centering
\includegraphics[width=\textwidth]{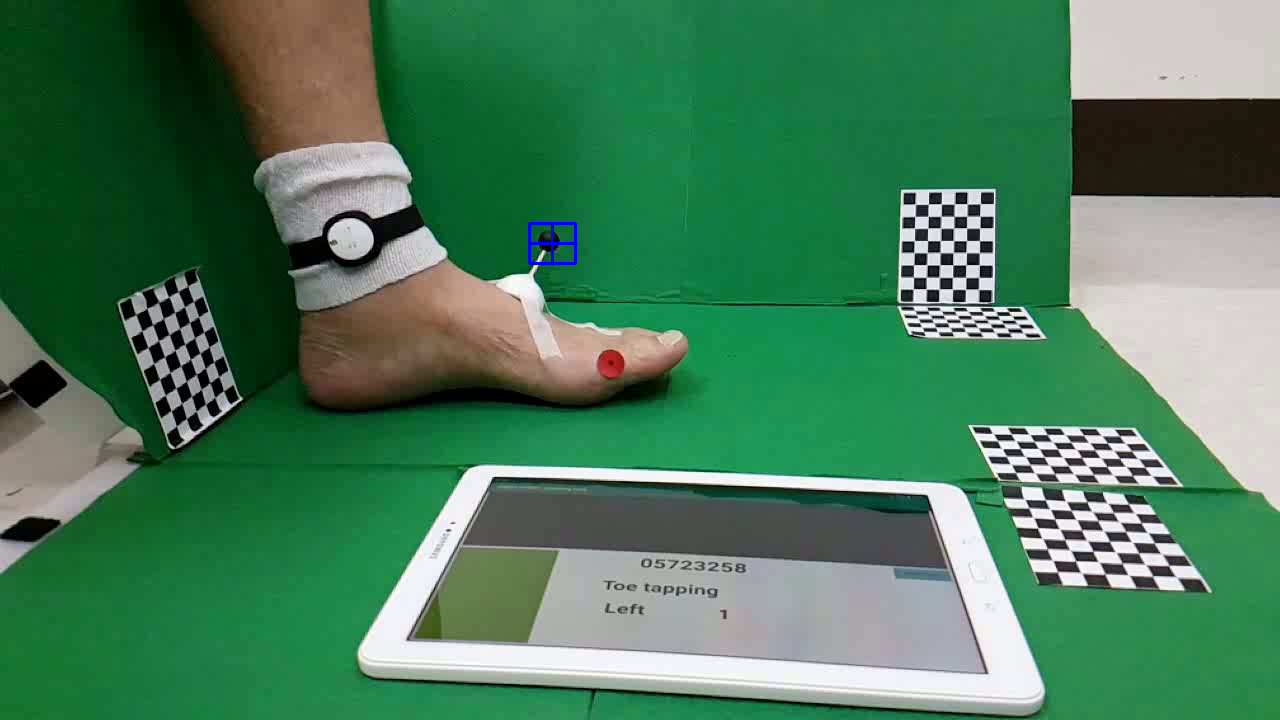}
\caption{Spherical marker extraction}
\label{fig:BlachSphereCrop}
\end{subfigure}
\caption{Extraction of background area (a) and black spherical marker (b).}
\label{fig:TwoCrop}
\end{figure}
Subsequently, any remaining green background regions surrounding the sphere were eliminated, and the 3C-FBI algorithm was employed to identify the circular structure corresponding to the black spherical marker.

\subsection{Ground Truth Labeling}
\begin{figure}[!htb]
\centering
\includegraphics[width=.115\textwidth]{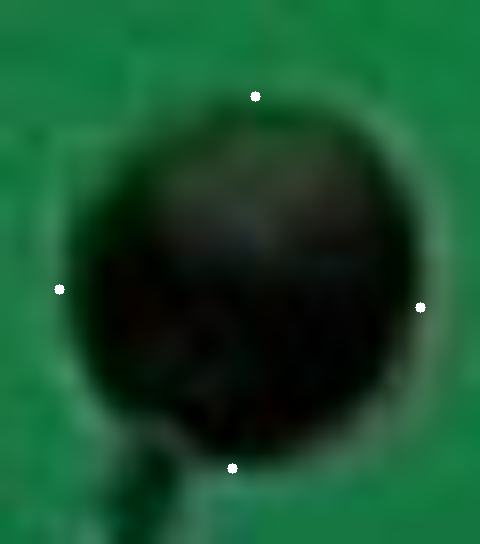}
\includegraphics[width=.115\textwidth]{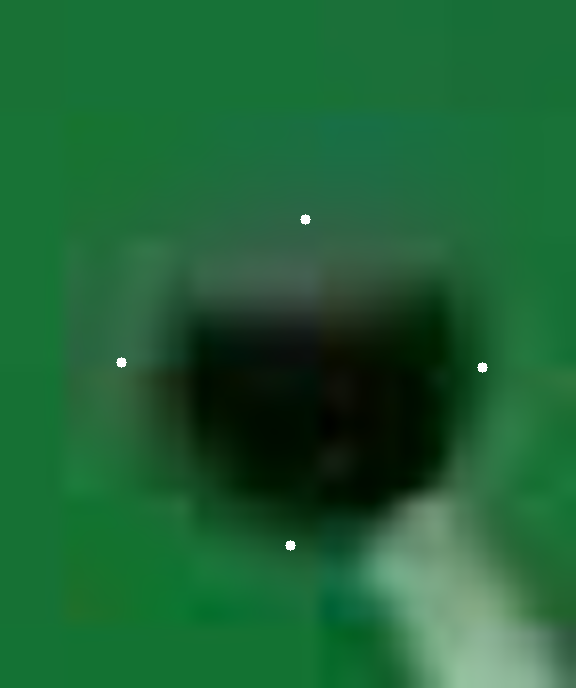}
\includegraphics[width=.115\textwidth]{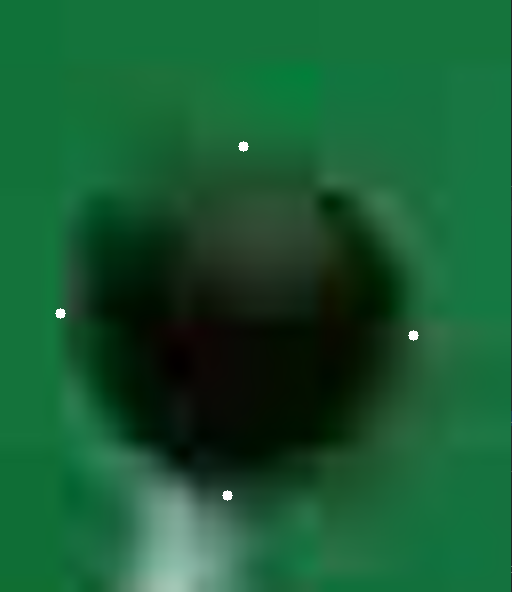}
\includegraphics[width=.115\textwidth]{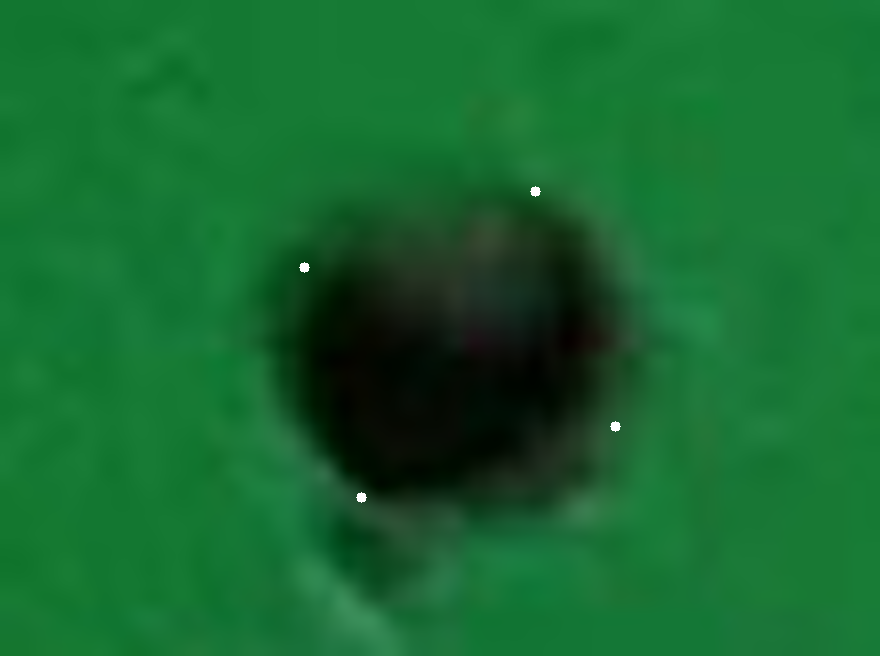}
\caption{Manual annotation outcomes for spherical markers in segmented images.}
\label{fig:spherelabeling}
\end{figure}

For validation of the 3C-FBI algorithm's performance, reference circles representing the black sphere were established through manual annotation.
Four perimeter points were manually identified on each sphere within every frame, from which circles were computed by selecting three points from the quartet, while the remaining point served to quantify annotation accuracy (Figure \ref{fig:spherelabeling}).

\begin{figure}[!htb]
    \centering
    \begin{subfigure}[b]{0.29\textwidth}
        \centering
        \includegraphics[width=\textwidth]{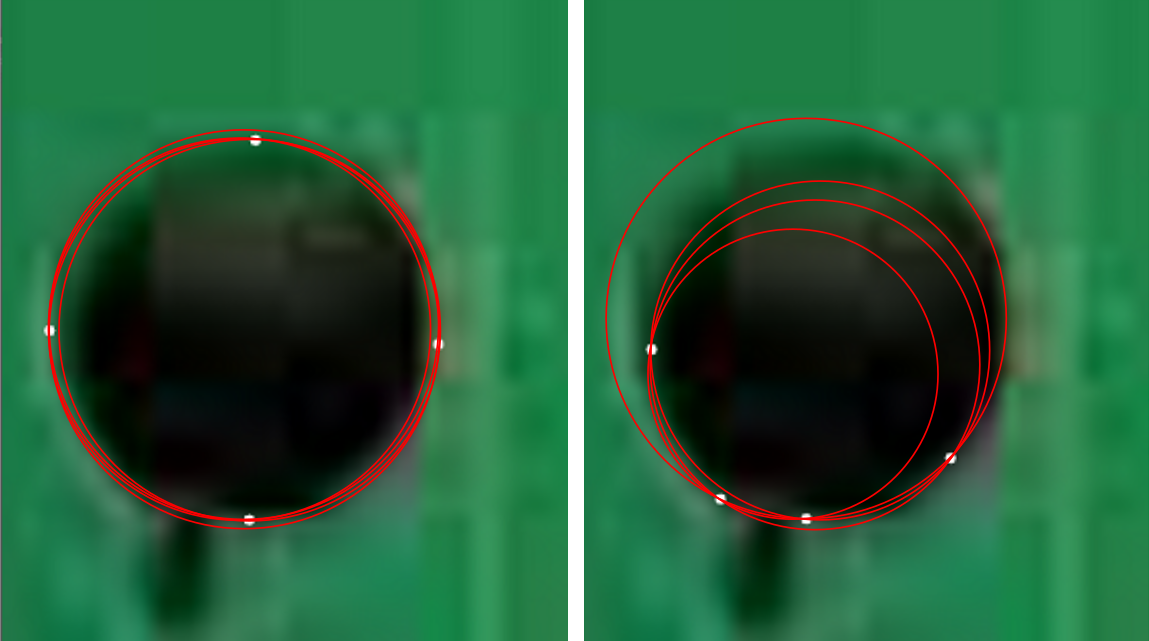}
        \caption{}
        \label{fig:circles}
    \end{subfigure}
    \hfill
    \begin{subfigure}[b]{0.19\textwidth}
        \centering
        \includegraphics[width=\textwidth]{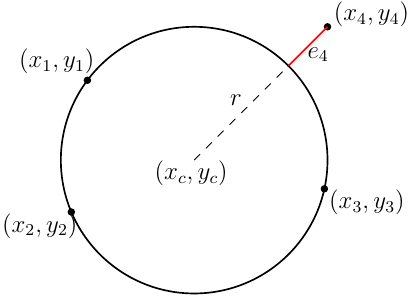}
        \caption{}
        \label{fig:CircleError}
    \end{subfigure}
    \caption{(a) Example of good (left) and poor (right) perimeter point selections. 
    (b) Annotation error $e_4$, calculated as the distance from point $(x_4, y_4)$ to the circle defined by points $(x_1, y_1)$, $(x_2, y_2)$, and $(x_3, y_3)$.}
    \label{fig:pointselection}
\end{figure}

To achieve sub-pixel precision, each segmented image underwent eight-fold magnification, expanding the sphere's radius from the original 9-14 pixels to 72-112 pixels.
Figure \ref{fig:pointselection} demonstrates the annotation error assessment by showing circles generated from both precise and imprecise point selections.

The annotation error for each point was quantified as its distance to circles constructed from the remaining three points, implementing a leave-one-out cross-validation methodology, as illustrated in Figure \ref{fig:CircleError}.

\subsection{Image Preprocessing}
The preprocessing pipeline for the 3C-FBI circle detection methodology encompassed evaluation of 19 distinct techniques to determine the most effective edge-image generation for optimal circle detection performance.
Two principal methodologies, GL (Green Level) and Med (Median Blurring), were implemented:
\begin{description}
\item[Group 1 (GL):] Nine image variants were generated without median filtering, employing Green Level thresholding within HSV color space to differentiate the black sphere from the green background environment.
\item[Group 2 (Med):] Ten filtered images were produced using variable median blur kernel sizes, subsequently processed through HSV transformation and green background masking similar to the GL approach.
\end{description}

\begin{figure*}[tb]
\centering
\includegraphics[width=0.7\textwidth]{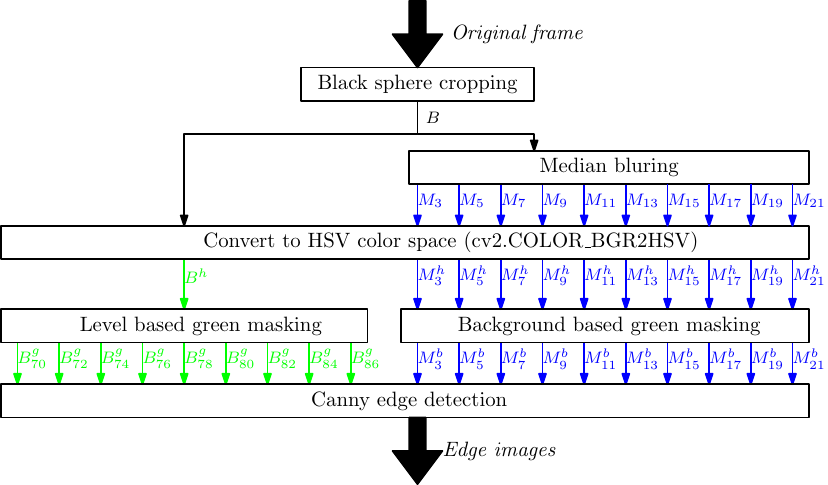}
\caption{Schematic representation of primary preprocessing workflow steps.}
\label{fig:CV_Tree3}
\end{figure*}
\begin{figure}[tb]
\centering
\begin{minipage}[t]{0.11\textwidth}
\centering
\includegraphics[width=1\textwidth]{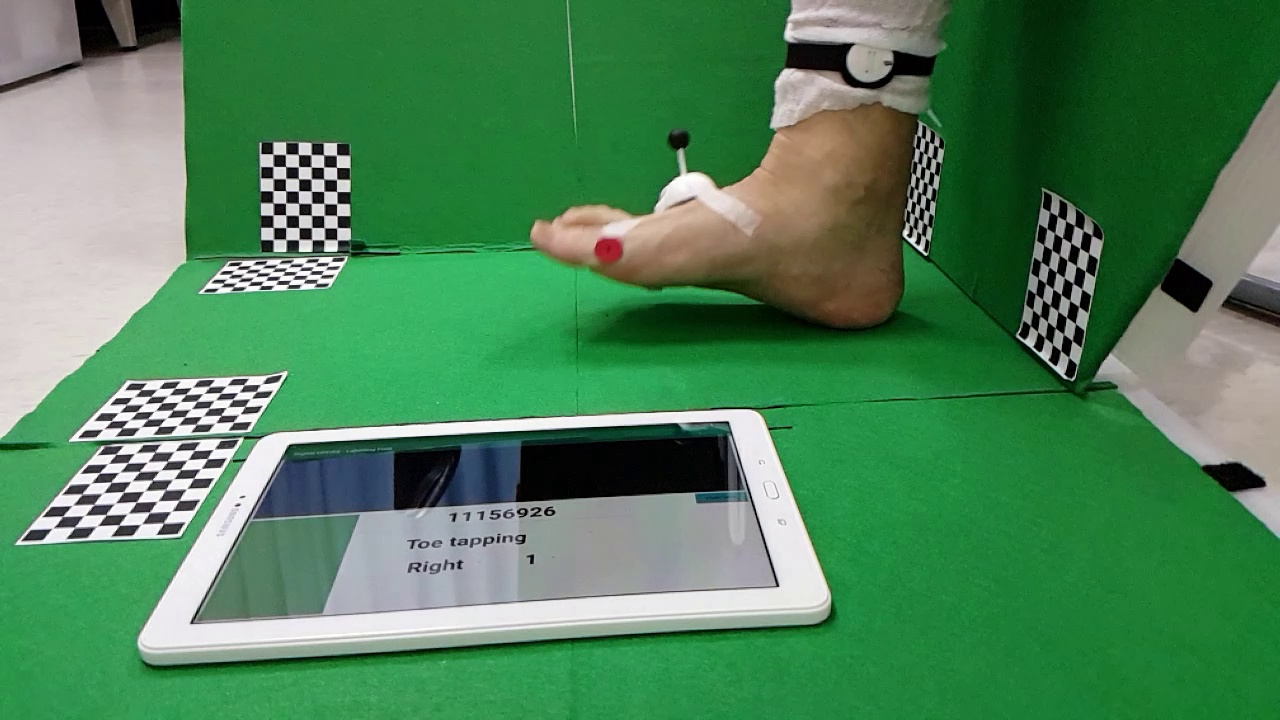}
\caption*{(a) Source image}
\end{minipage}\hfill
\begin{minipage}[t]{0.11\textwidth}
\centering
\includegraphics[width=1\textwidth]{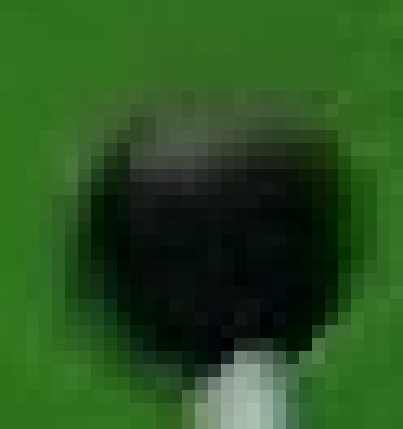}
\caption*{(b) Spherical marker ROI}
\end{minipage}
\hfill
\begin{minipage}[t]{0.11\textwidth}
\centering
\includegraphics[width=1\textwidth]{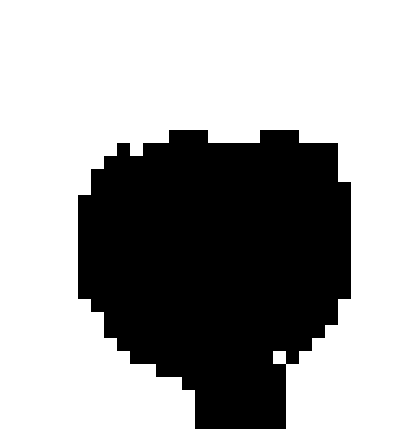}
\caption*{(c) Background mask}
\end{minipage}
\hfill
\begin{minipage}[t]{0.11\textwidth}
\centering
\includegraphics[width=1\textwidth]{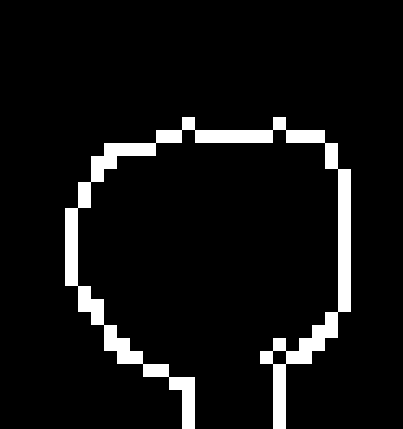}
\caption*{(d) Edge feature extraction}
\end{minipage}
\caption{Sequential preprocessing operations: \textbf{(a) Source image} represents the raw frame captured from the video stream, \textbf{(b) Spherical marker ROI} identifies the region containing the black spherical marker, \textbf{(c) Background mask} shows the mask application for green component isolation, and \textbf{(d) Edge feature extraction} illustrates the detected edges within the masked region.}
\label{fig:preprocessing}
\end{figure}

All 19 binary images underwent Canny edge detection to generate edge-enhanced images suitable for circle detection analysis (Figure \ref{fig:CV_Tree3}).
The sequential preprocessing stages are visualized in Figure \ref{fig:preprocessing}.

\subsection{Combinatorial Complexity}

An edge-enhanced image containing \( N \) pixels yields \( \binom{N}{3} \) possible triplet combinations, underscoring the significant computational complexity associated with circle detection tasks.

\begin{figure}[tb]
	\centering
	\includegraphics[width=0.48\textwidth]{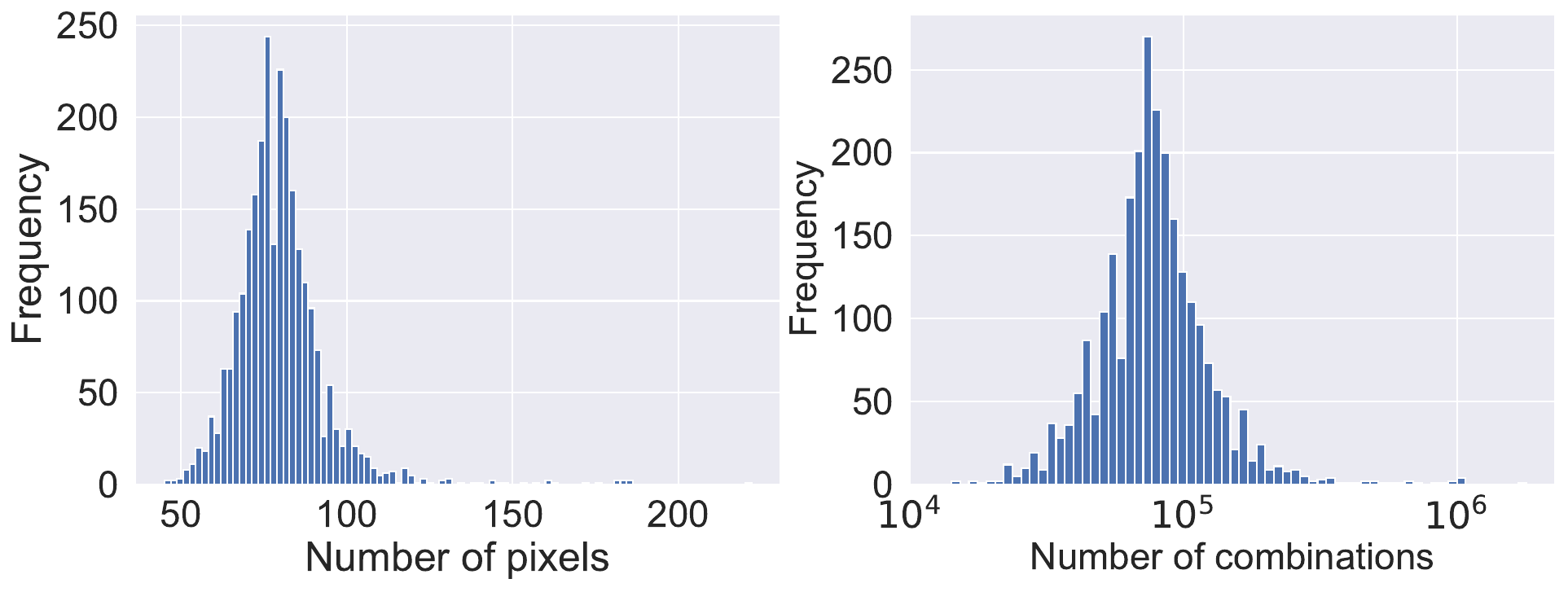}
	\caption{Histogram of the number of edgels per frame (left) and histogram of the number of combinations per frame on a $log_{10}$ scale (right).}
	\label{fig:PixelCount}
\end{figure}

Figure~\ref{fig:PixelCount} shows a histogram of the distribution of edgels and corresponding combination counts across the 2,736 processed frames (19 preprocessing variants × 144 original frames).

To reduce computational load, the 3C-FBI algorithm uses a fixed number \( N_{\text{triplets}} \) of randomly selected triplets per image. We evaluated values ranging from 500 to 10,000 and observed optimal performance at 5,000 triplets.

\subsection{A Combinatorial Approach using Convolutions for Circle Fitting in Blurry Images (3C-FBI)}

The 3C-FBI algorithm is designed to robustly identify circular structures in edge-detected images, retaining high performance even under challenging conditions such as noise, blur, and partial occlusion.

This method integrates two key strategies: combinatorial sampling of edgels and convolution-based refinement in parameter space. The algorithm begins by randomly selecting a fixed number of triplets from the detected edgels set. Each triplet uniquely defines a circle based on geometric constraints, enabling the generation of a diverse set of circle candidates from a single frame.

These candidates are then projected onto a discretized three-dimensional accumulator space—analogous to the Hough Transform \cite{duda1972use}—which represents potential center coordinates \((x, y)\) and radii \(r\). To reinforce consensus and suppress noise, the algorithm applies localized convolution operations over this accumulator, smoothing the vote distribution and highlighting dense regions that correspond to likely circle parameters.

Final parameter estimation is performed by identifying the dominant peaks in the smoothed accumulator and refining them using a center-of-mass calculation.

\begin{figure*}[!htbp]
    \centering
    \begin{minipage}[t]{0.22\textwidth}
        \centering
        \includegraphics[width=1\textwidth]{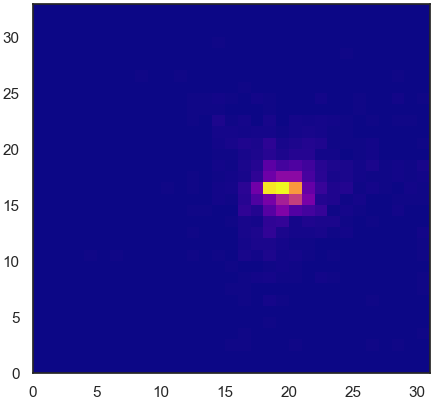}
        \caption*{(e) Center votes}
    \end{minipage}
    \hfill
    \begin{minipage}[t]{0.22\textwidth}
        \centering
        \includegraphics[width=1\textwidth]{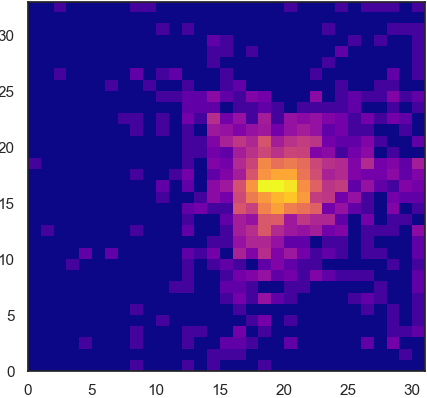}
        \caption*{(f) Center votes $\log(\text{votes} + 1)$}
    \end{minipage}
    \hfill
    \begin{minipage}[t]{0.22\textwidth}
        \centering
        \includegraphics[width=1\textwidth]{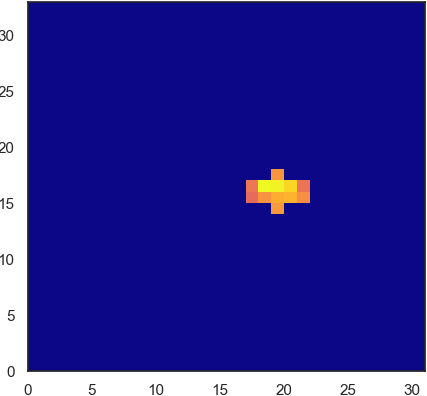}
        \caption*{(g) Most voted centers}
    \end{minipage}
    \hfill
    \begin{minipage}[t]{0.22\textwidth}
        \centering
        \includegraphics[width=1\textwidth]{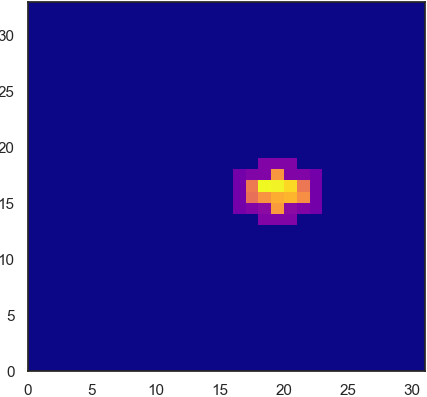}
        \caption*{(h) Neighborhood search }
    \end{minipage}
    \caption{Results of the 3C-FBI method: \textbf{(e) Center votes} show the initial detection of all possible circle centers within the image, \textbf{(f) Center votes $\log(\text{votes} + 1)$} applies a logarithmic transformation to enhance the visibility of the center points, \textbf{(g) Most voted centers} highlights the top likely circle centers based on the voting mechanism, and \textbf{(h) Neighborhood search} focuses on the local neighborhood around the top detected centers, refining the circle’s position and radius.}
    \label{fig:3C-FBI}
\end{figure*}

The pseudocode for the full algorithm is provided in Algorithm~\ref{alg:main}, and the final four stages of the detection process are illustrated in the heatmaps shown in Figure~\ref{fig:3C-FBI}.

This hybrid framework allows 3C-FBI to maintain precision and robustness, making it especially effective for real-time applications and scenarios with degraded image quality. The graphical abstract outlines the entire 3C-FBI pipeline, highlighting its major stages of detection and parameter refinement.

\begin{algorithm}[!ht]
\caption{3C-FBI: Combinatorial Convolution-based Circle Fitting in Blurry Images}
\label{alg:main}
\KwIn{Edge-detected image}
\KwOut{Estimated circle center $(x_c, y_c)$ and radius $r$}

\textbf{Step 1: Load Edge-detected image};

Load a binary image where edgels are black and the rest white (from methods such as Canny or HSV-based preprocessing).

\textbf{Step 2: Random Triplet Sampling};

Randomly sample $N_{\text{triplets}}$ (5000) triplets of edgels;
 
\ForEach{triplet $(p_1, p_2, p_3)$}{
    Compute the circle $(x, y, r)$ passing through the 3 points (Eq. \ref{eq:circle_formula});
    Accumulate a vote in a discretized $(x, y, r)$ accumulator array
}

\textbf{Step 3: Circle Candidate Selection};

Select the top N  ($5$) most voted accumulator coordinates;

\textbf{Step 4: Convolutional Refinement};

\ForEach{of the N most voted accumulator coordinates}{
Apply a convolution using the infinity norm as kernel with radius $r_\infty$ ($1$) over the accumulator to smooth the number of votes;
}
\textbf{Step 5: Final Estimation via Center of Mass};

Compute the weighted center of mass over the most voted accumulator coordinate and its 26 neighbors to refine $(x_c, y_c, r)$;

\Return $(x_c, y_c, r)$

\end{algorithm}

\subsection{Computations}

The computations were performed on an MSI GE75 laptop, equipped with an Intel i7-10875H CPU, an NVIDIA RTX 2070 GPU, and 64 GB of RAM.
 \section{Results}
\label{sec:Results}

Our evaluation consists of three distinct experiments designed to thoroughly assess circle-fitting performance. The first experiment (A) utilizes real-world data from quantitative PD assessments, leveraging video sequences from 18 subjects performing toe-tapping tests. The second set of experiments (B) employs controlled synthetic data in two configurations:
\begin{itemize}
    \item B1: A comparative analysis following the experimental framework of \citet{qi2024robust}, using points distributed along a semicircle
    \item B2: An extended evaluation using complete circles at varying point resolutions to assess method robustness
\end{itemize}
Across all experiments, we compare our proposed method against seven established circle-fitting approaches: RHT \citep{xu1990new}, RCD \citep{chen2001efficient}, RFCA \citep{ladron2011robust}, \citet{nurunnabi2018robust}, and more recent methods including \citet{guo2019iterative}, \citet{greco2023impartial}, and \citet{qi2024robust}. This set is representative both for classical and the latest circle fitting methods. This structure provides a comprehensive evaluation framework spanning both practical applications and controlled theoretical scenarios.

\subsection{Experiment A: Real-World data from Parkinson's Disease assesment}

In our analysis of real-world data from the toe-tapping test, which involved video sequences from 18 subjects (12 PD, 6 controls), selecting an effective preprocessing method was crucial for optimal circle detection. GL thresholding showed stability across its range, with thresholds from 70 to 86 maintaining consistently high mean Jaccard Index values above 0.85. The top three performing GL thresholds were GL84 (0.877), GL82 (0.877), and GL80 (0.877), achieving nearly identical performance. These three thresholds were selected for further analysis due to their optimal performance and their consecutive nature, suggesting a stable optimal region for the GL parameter. This stability in the 80-84 range indicates a robust preprocessing window that effectively isolates the black sphere from the green background.

Median Blur was more sensitive to parameters. While moderate kernel sizes (Med7-Med9) achieved respectable performance with Jaccard indices around 0.86, larger kernels caused a significant decline in accuracy, dropping to 0.673 for Med19, due to excessive smoothing. This degradation in performance with larger kernel sizes suggests that Median Blur, while effective in moderate ranges, is less reliable than GL thresholding for this application.

\begin{figure*}[tb]
\begin{center}
    \centering
    \makebox[\textwidth][c]{\includegraphics[width=1\textwidth]{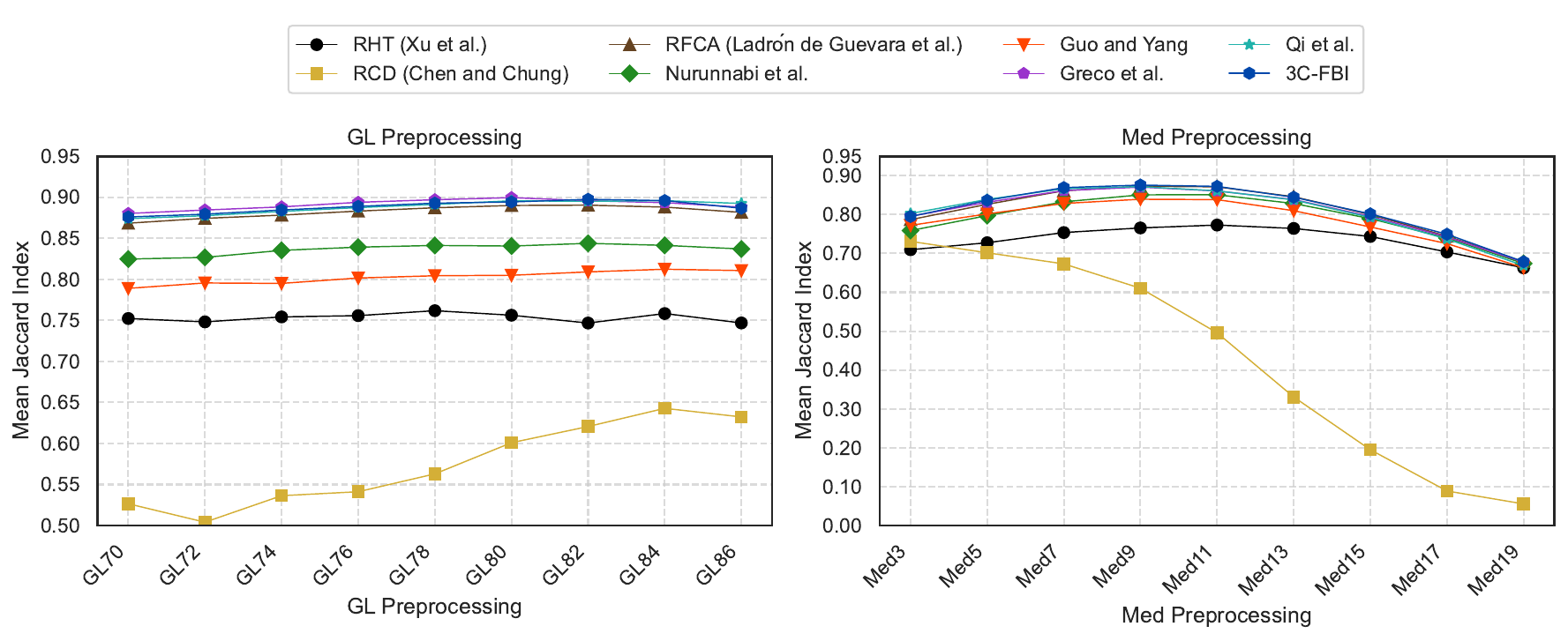}}
    \caption{Comparative analysis of circle detection methods across different preprocessing techniques. Left: Performance under Green Level (GL) thresholding preprocessing, showing consistent high performance (Jaccard Index 0.84-0.90) across the top three methods, with \citet{guo2019iterative} achieving slightly superior results. Right: Performance under Median Blur (Med) preprocessing, demonstrating a clear degradation in performance with increasing kernel size for all methods after Med11. The Jaccard Index ranges from 0.70 to 0.85, indicating that GL preprocessing generally yields more robust results than Median Blur for all evaluated methods.}
    \label{fig:Roman2024_5_Jaccard}
\end{center}
\end{figure*}

Based on these results, we analyze the performance using the three optimal GL preprocessing thresholds (GL80, GL82, and GL84). Table~\ref{tab:bestGL} presents the mean performance metrics across these thresholds for each method.

\begin{table*}[htbp]
\centering
\caption{Mean values of quantitative comparison of circle detection methods using three optimal Green Level preprocessing thresholds (GL80, GL82 and GL84). Performance metrics include Jaccard Index (higher is better, range 0-1), Average Distance (AD) in pixels (lower is better), Root Mean Square Error (RMSE) in pixels (lower is better), and computation time in frames per second (FPS), escluding preprocessing. Results show that modern methods (\textbf{\citet{greco2023impartial}}, \textbf{\citet{qi2024robust}} and \textbf{3C-FBI}) achieve comparable high performance (Jaccard $\approx$ 0.9, AD $\approx$ 2.4 px, RMSE $\approx$ 1.1 px), significantly outperforming classical approaches (RHT and RCD). Values represent averages across all 144 test images.}
\label{tab:bestGL}
\begin{tabular}{lcccc}
\hline
Method & Jaccard & AD & RMSE & FPS\\
\hline
RHT (\citet{xu1990new}) & 0.757 & 3.539 & 1.694 & 26.8\\
RCD (\citet{chen2001efficient}) & 0.620 & 7.318 & 4.536 &  39.5\\
RFCA (\citet{ladron2011robust}) & 0.889 & 2.467 & 1.108 & 91.1\\
\citet{nurunnabi2018robust} & 0.843 & 2.814 & 1.237 & 236.1\\
\citet{guo2019iterative} & 0.809 & 3.232 & 1.358 & 2700\\
\textbf{\citet{greco2023impartial}} & 0.896 & 2.370 & 1.082 & 111.4\\
\textbf{\citet{qi2024robust}} & 0.896 & 2.353 & 1.089 & 106.9\\
\textbf{3C-FBI} & 0.896 & 2.388 & 1.083 & 69.3\\
\hline
\end{tabular}
\end{table*}

\subsection{Experiment B: Synthetic Data Evaluation}

We conducted controlled experiments with artificial data to systematically evaluate algorithm performance under known ground truth conditions in a manner comparable to previously reported results.

\subsubsection{Configuration B1:}
\begin{figure*}[!htbp]
\begin{center}
    \centering
    \makebox[\textwidth][c]{\includegraphics[width=1.2\textwidth]{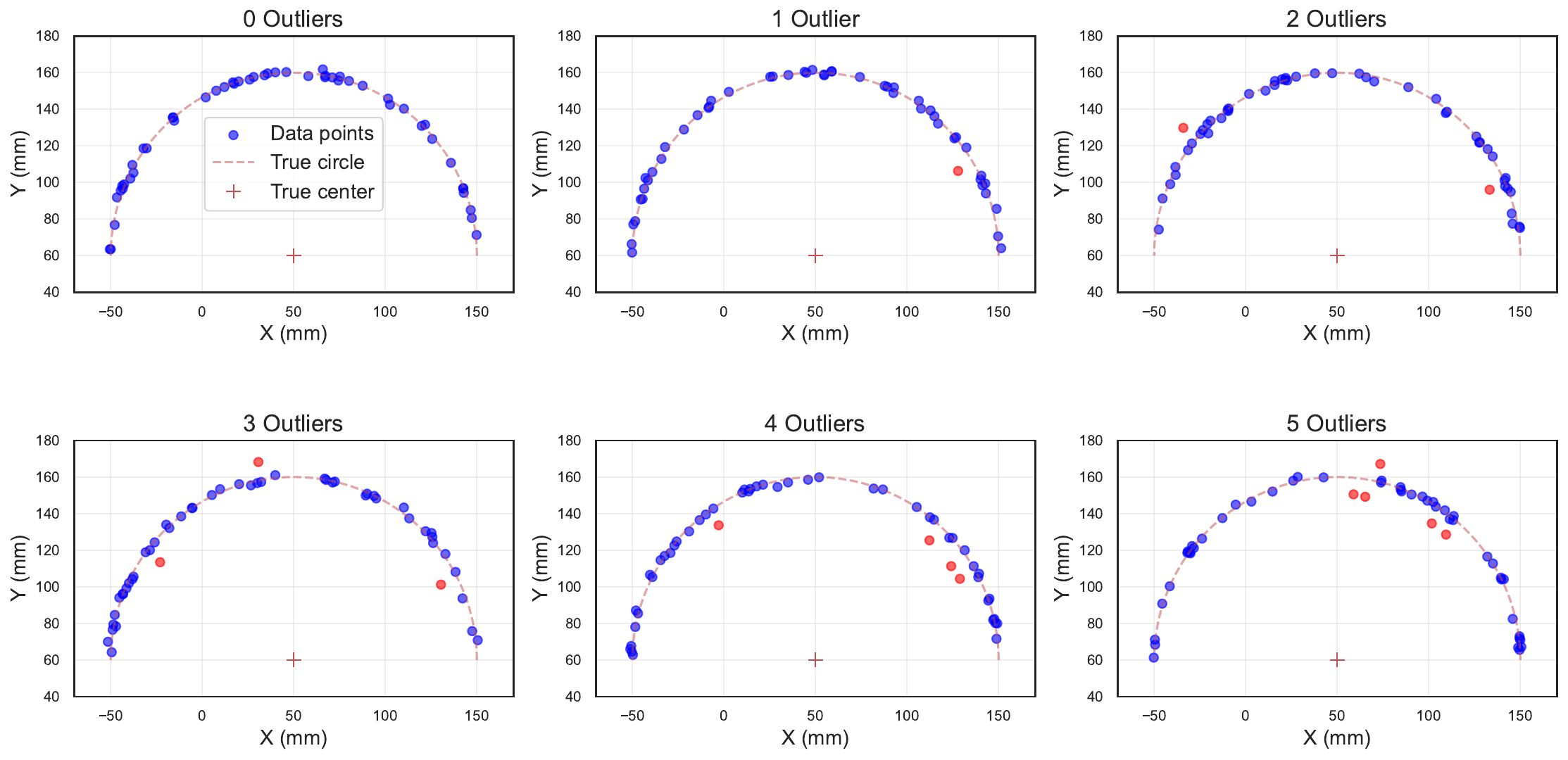}}
    \caption{Illustration of synthetic datasets with varying numbers of outliers (0-5). Blue points represent inliers, red points indicate outliers, and the red dashed line shows the true circle with center (50, 60) mm and radius 100 mm. The datasets consist of 50 points randomly distributed along the upper semicircle with Gaussian noise ($\sigma = 1$ mm) added to the radial direction.}
    \label{fig:Fig23}
\end{center}
\end{figure*}

Following the framework established by \citet{qi2024robust}, points were distributed along a semicircle with center $({\tilde{x}}_0, {\tilde{y}}_0) = (50, 60)$ mm and radius ${\tilde{r}}_0 = 100$ mm. As shown in Figure~\ref{fig:Fig23}, Gaussian noise ($\sigma = 1$ mm) was added to simulate measurement errors, and outliers with magnitudes of $5\sigma$ to $10\sigma$ were introduced in varying quantities (0-5 points) to test robustness. 

The experimental results, presented in Tables~\ref{tab:jaccard_results}, \ref{tab:AD_results}, and \ref{tab:RMSE_results}, demonstrate the comparative performance of various circle-fitting methods. The Jaccard index analysis shows that all methods, except RHT maintain consistently high accuracy ($>0.98$) regardless of outlier count. Notably, \citet{qi2024robust}'s method and RFCA achieves the highest mean Jaccard index (0.991), closely followed by our 3C-FBI (0.989). This is also reflected in AD and RMSE, albeit \citet{qi2024robust} being the winner.

\begin{table*}[htbp]
\centering
\caption{Jaccard Index when tested on semi circles under varying outlier conditions}
\label{tab:jaccard_results}
\makebox[\textwidth][c]{\begin{tabular}{lccccccc}
\hline
Method \textbackslash $\;$No. outliers & 0 & 1 & 2 & 3 & 4 & 5 & Mean \\
\hline
RHT - \citet{xu1990new}  			& 0.973 & 0.967 & 0.964 & 0.958 & 0.961 & 0.959 & 0.964 \\
RCD - \citet{chen2001efficient} 		& 0.986 & 0.988 & 0.986 & 0.987 & 0.987 & 0.986 & 0.987 \\
RFCA - \citet{ladron2011robust} 	& \textbf{0.991} &\textbf{ 0.991} & \textbf{0.991} & \textbf{0.991} & 0.991 & \textbf{0.991} & \textbf{0.991} \\
\citet{nurunnabi2018robust} 		& \textbf{0.991} & 0.987 & 0.984 & 0.983 & 0.982 & 0.980 & 0.984 \\
\citet{guo2019iterative} 				& 0.989 & 0.987 & 0.988 & 0.988 & 0.985 & 0.984 & 0.987 \\
\textbf{\citet{greco2023impartial}} 	& 0.988 & 0.988 & 0.988 & 0.989 & 0.989 & 0.988 & 0.988 \\
\textbf{\citet{qi2024robust}} 		& \textbf{0.991} & \textbf{0.991} & \textbf{0.991} & \textbf{0.991 }& \textbf{0.992} & \textbf{0.99}1 & \textbf{0.991} \\
\textbf{3C-FBI} 						& 0.989 & 0.990 & 0.989 & 0.988 & 0.988 & 0.989 & 0.989 \\
\hline
\end{tabular}}
\end{table*}

\begin{table*}[htbp]
\centering
\caption{AD when tested on semi circles  of different circle fitting methods (unit: mm)}
\label{tab:AD_results}
\makebox[\textwidth][c]{\begin{tabular}{lccccccc}
\hline
Method \textbackslash $\;$No. outliers & 0 & 1 & 2 & 3 & 4 & 5 & Mean \\
\hline
RHT - \citet{xu1990new}  & 1.735 & 2.098 & 2.355 & 2.696 & 2.485 & 2.682 & 2.342 \\
RCD - \citet{chen2001efficient} & 0.899 & 0.790 & 0.882 & 0.785 & 0.817 & 0.884 & 0.843 \\
RFCA - \citet{ladron2011robust} & 0.522 & 0.557 & 0.593 & 0.550 & 0.569 & 0.583 & 0.562 \\
\citet{nurunnabi2018robust} & 0.561 & 0.844 & 1.013 & 1.022 & 1.177 & 1.302 & 0.987 \\
\citet{guo2019iterative} & 0.714 & 0.799 & 0.763 & 0.718 & 0.905 & 1.009 & 0.818 \\
\textbf{\citet{greco2023impartial}} & 0.743 & 0.777 & 0.756 & 0.692 & 0.713 & 0.729 & 0.735 \\
\textbf{\citet{qi2024robust}} & 0.593 & 0.536 & 0.560 & 0.562 & 0.532 & 0.536 & 0.553 \\
\textbf{3C-FBI} & 0.685 & 0.658 & 0.703 & 0.764 & 0.764 & 0.682 & 0.709 \\
\hline
\end{tabular}}
\end{table*}

\begin{table*}[htbp]
\centering
\caption{RMSE when tested on semi circles  of different circle fitting methods (unit: mm)}
\label{tab:RMSE_results}
\makebox[\textwidth][c]{\begin{tabular}{lccccccc}
\hline
Method \textbackslash $\;$No. outliers & 0 & 1 & 2 & 3 & 4 & 5 & Mean \\
\hline
RHT - \citet{xu1990new} & 1.220 & 1.653 & 1.719 & 2.012 & 1.946 & 2.058 & 1.768 \\
RCD - \citet{chen2001efficient} & 0.625 & 0.577 & 0.637 & 0.613 & 0.608 & 0.655 & 0.619 \\
RFCA - \citet{ladron2011robust} & 0.397 & 0.475 & 0.467 & 0.426 & 0.434 & 0.465 & 0.444 \\
\citet{nurunnabi2018robust} & 0.456 & 0.659 & 0.755 & 0.842 & 0.839 & 0.989 & 0.757 \\
\citet{guo2019iterative} & 0.571 & 0.621 & 0.602 & 0.598 & 1.114 & 1.254 & 0.793 \\
\textbf{\citet{greco2023impartial}} & 0.550 & 0.592 & 0.587 & 0.553 & 0.519 & 0.586 & 0.564 \\
\textbf{\citet{qi2024robust}} & 0.420 & 0.421 & 0.503 & 0.414 & 0.394 & 0.427 & 0.430 \\
\textbf{3C-FBI} & 0.492 & 0.471 & 0.550 & 0.573 & 0.573 & 0.573 & 0.539 \\
\hline
\end{tabular}}
\end{table*}

\subsubsection{Configuration B2:}

To systematically evaluate circle-fitting performance, we developed a comprehensive experimental framework centered on a reference circle with center $({\tilde{x}}_0, {\tilde{y}}_0) = (120, 120)$ mm and radius ${\tilde{r}}_0 = 120$ mm. The framework generates $N=100$ points along this circle, with their positions modified according to three key parameters: measurement uncertainty, outlier contamination, and spatial discretization.

The measurement uncertainty was implemented as additive Gaussian noise $\mathcal{N}(0,\sigma^2)$ applied radially to each point, where $\sigma$ was defined as a percentage of the true circle radius $r_0$. Specifically, noise levels of $\{0\%, 1\%, 2\%, 5\%, 10\%\}$ of $r_0 = 120\,\text{mm}$ were used, corresponding to standard deviations $\sigma \in \{0, 1.2, 2.4, 6.0, 12.0\}\,\text{mm}$. The points were initially defined in polar coordinates $(r,\theta)$, with $r = r_0$ and $\theta$ uniformly distributed around the circle. We note that angular noise was not considered as the uniform distribution of $N=100$ points around the circle already provides sufficient angular variation. Preliminary experiments showed no significant impact of additional angular perturbations on the results.

For outlier analysis, we randomly replaced proportions $p \in \{0\%, 10\%, 20\%, 50\%, 75\%\}$ of circle points with outliers uniformly distributed within a square region of side length $4r_0$ centered at $(x_0, y_0)$. 
Extending beyond previous work, we implemented a minimum separation criterion requiring each outlier's distance from the true circle to exceed $3\sigma$, ensuring differentiation between outliers and noise-affected inlier points, which—with 99.7\% probability under Gaussian noise—lie within three standard deviations.

To evaluate spatial discretization effects, we introduced a quantization parameter $q \in  \{0, 1, 2, 3, 6, 12, 24, 40\}$ defining the spatial sampling granularity. For $q = 0$, points maintain their continuous coordinates, while for $q > 0$ coordinates undergo quantization via:

$$ (x, y, r) \rightarrow \left(\text{round}\left(\frac{x}{q}\right), \text{round}\left(\frac{y}{q}\right), \text{round}\left(\frac{r}{q}\right)\right) $$

\noindent where $(x, y)$ represents spatial coordinates and $r$ denotes the radius.

Image resolution $R_q$ was systematically defined as a function of the quantization parameter, where $R_0$ represents infinite resolution (continuous domain), and $R_1 = 480 \times 480$ pixels serves as our baseline discrete resolution. For arbitrary quantization $q > 1$, the resolution $R_q$ is given by:

$$R_q = \left(\frac{480}{q}\right) \times \left(\frac{480}{q}\right)$$

\noindent with the circle center coordinates scaled proportionally as:

$$(x_c, y_c)_q = \left(\frac{240}{q}, \frac{240}{q}\right)$$

\begin{figure*}[!htbp]
\begin{center}
    \centering
    \makebox[\textwidth][c]{\includegraphics[width=\textwidth]{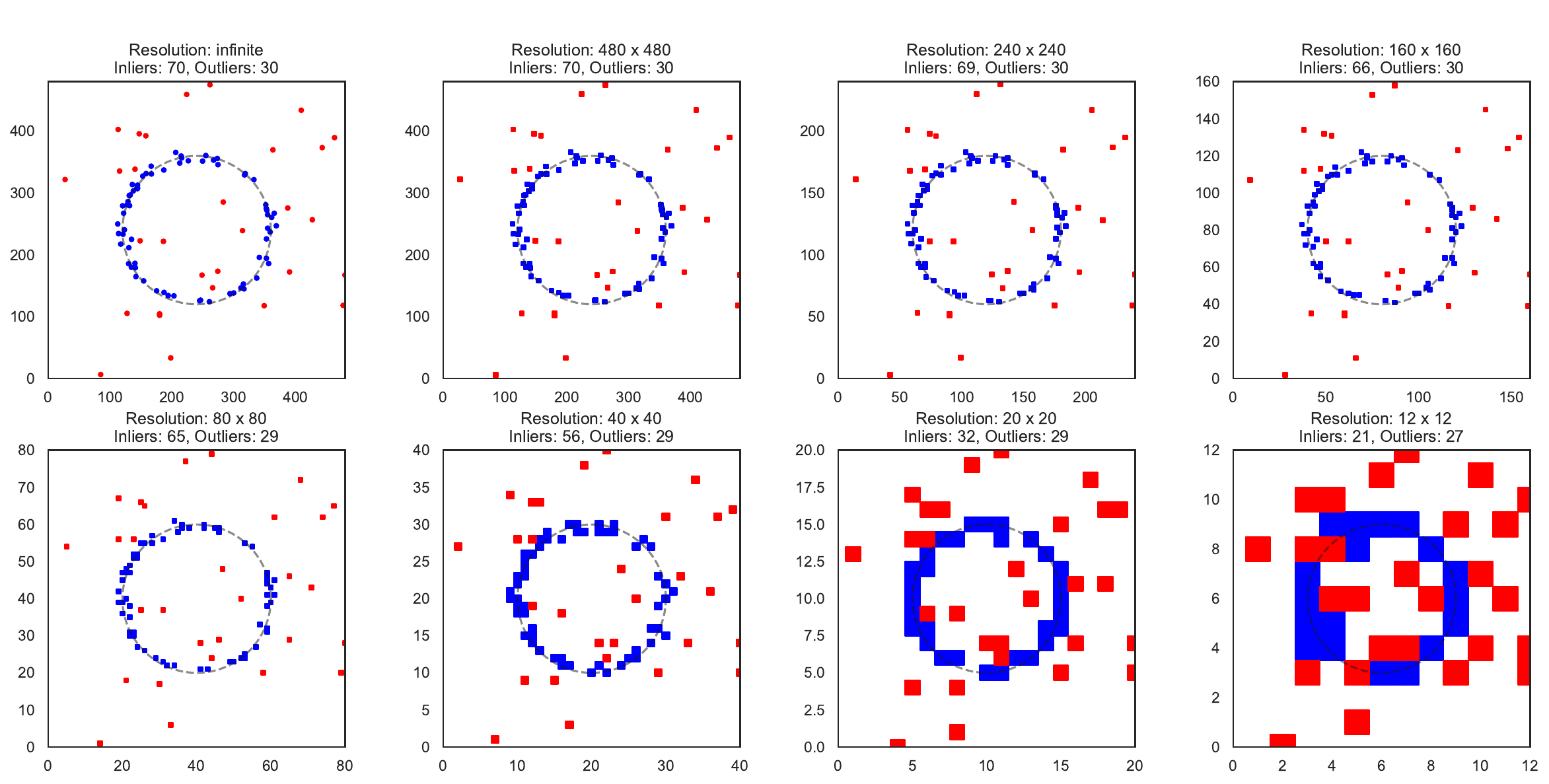}}
    \caption{Visualization of spatial quantization effects using a test dataset generated from a circle with center (120,120) mm and radius 120 mm, containing 100 points (70 inliers, 30 outliers). The sequence shows progressive coordinate discretization through increasing partition scales q, from continuous coordinates (infinite resolution) to coarse discretization (12×12 pixels). Blue points represent inliers identified along the reference circle (shown by connecting lines), while red points indicate outliers uniformly distributed in the bounding region. The sequence demonstrates how spatial quantization impacts the visibility and structure of the circular pattern.}
    \label{fig:Fig24}
\end{center}
\end{figure*}

This quantization effectively partitions the coordinate space into a $\frac{480}{q}$-sized grid, with all points in each cell mapped to the cell center. Increasing $q$ produces progressively coarser discretization, enabling systematic evaluation of algorithm robustness to resolution effects. Figure~\ref{fig:Fig24} shows an example of how lowering the resolution affects the image. For all cases, we conducted $100$ independent trials per configuration and report the mean to ensure statistical significance.

\begin{figure*}[!htbp]
\begin{center}
    \centering
    \makebox[\textwidth][c]{\includegraphics[width=1.2\textwidth]{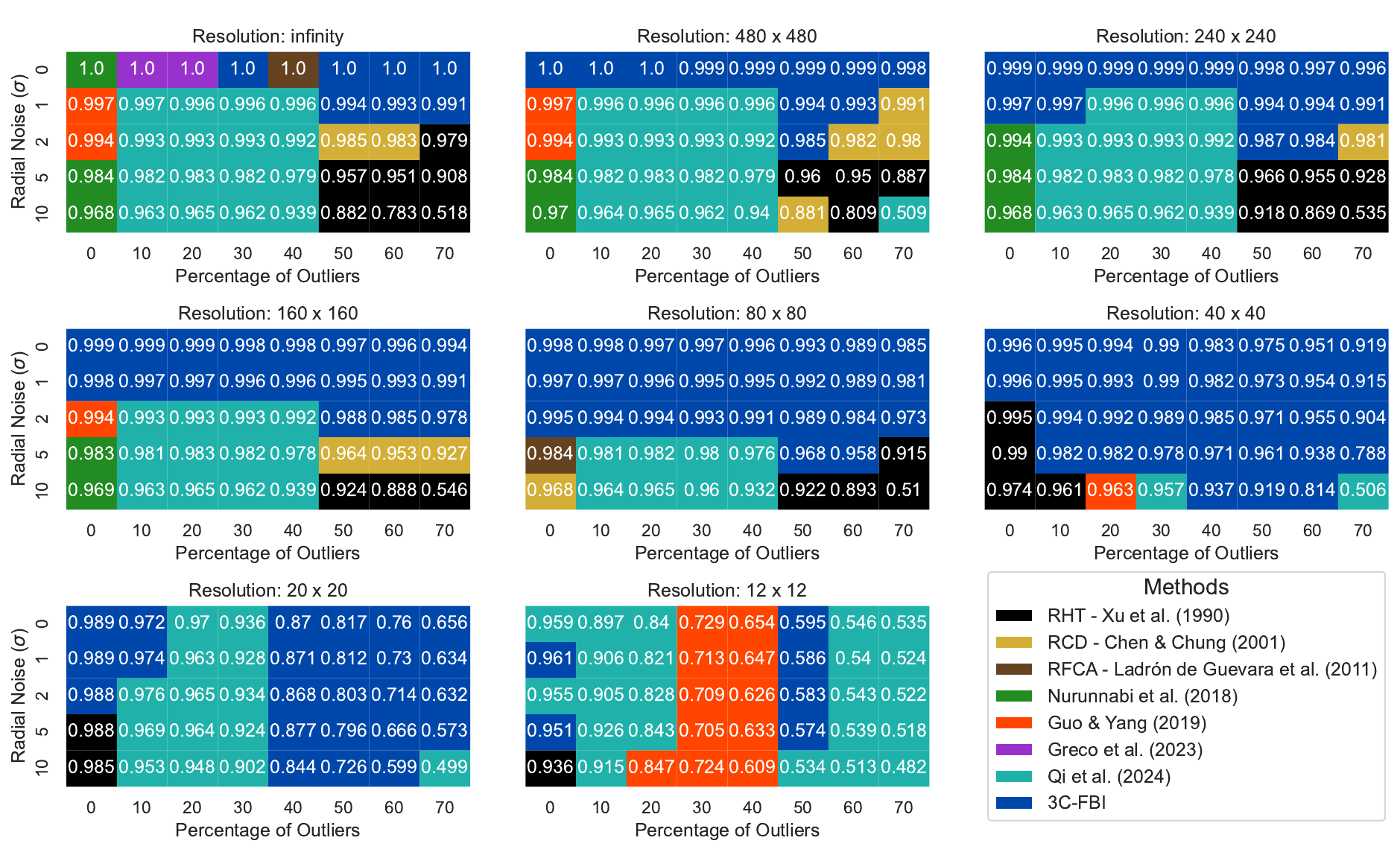}}
    \caption{The best performing circle fitting method (highest Jaccard index) is highlighted for each specific combination of resolution, radial noise and outlier contamination. The mean across 100 realizations is shown.}
    \label{fig:Roman2024_Jaccard_Heatmap_mean}
\end{center}
\end{figure*}

The comparative performance analysis illustrated in Figure \ref{fig:Roman2024_Jaccard_Heatmap_mean} reveals distinct patterns of algorithmic superiority across varying experimental configurations. This heatmap matrix presents the best-performing methods for each combination of resolution (ranging from infinite to $12 \times 12$ pixels), noise ratio ($0-10\%$), and outlier contamination ($0-75\%$). 

Table \ref{tab:WinnerCount} quantifies these results by tallying the number of configurations in which each method achieved the highest Jaccard Index. Notably, the 3C-FBI algorithm demonstrates superior performance in 44.1\% (141/320) of the tested configurations, establishing it as the dominant approach overall. This is followed by \citet{qi2024robust} method, which leads in 33.1\% (106/320) of cases. The remaining six methods collectively account for only 22.8\% of winning configurations, with no individual method exceeding 10\%.

Further analysis reveals that 3C-FBI exhibits particular strength at moderate resolutions, winning in 20 configurations at $160 \times 160$ pixels (50\%), 27 at $80 \times 80$ pixels (67.5\%), and 33 at $40 \times 40$ pixels (82.5\%). This suggests exceptional robustness to moderate spatial discretization effects. In contrast, \citet{qi2024robust} method dominates at extreme resolutions, leading in 17 configurations (42.5\%) at infinite resolution and 22 configurations (55\%) at the lowest $12 \times 12$ resolution.

The Jaccard Index results demonstrate that circle detection performance achieves near-perfect accuracy ($\sim 1.0$) under optimal conditions at high resolutions ($480 \times 480$ pixels and infinite) with minimal noise and outliers. The best algorithms maintain robust performance (Jaccard Index $>0.95$) down to $160 \times 160$ pixel resolution with up to 20\% outlier contamination. At the highest resolutions, this performance stability persists even with 20\% outliers present. However, a significant performance degradation occurs as resolution decreases below $80 \times 80$ pixels, becoming particularly pronounced in the $40 \times 40$ to $12 \times 12$ range. This degradation is most evident at $12 \times 12$ resolution, where the Jaccard index falls below $0.7$ even under minimal noise and outlier conditions, indicating that spatial discretization becomes a fundamental limiting factor at such coarse resolutions. These findings establish clear operational boundaries for reliable circle detection, with optimal performance requiring resolutions of at least $160 \times 160$ pixels and moderate noise ($\leq 2\%$) and outlier ($\leq 20\%$) levels.

\begin{table*}[htbp]
\footnotesize
\caption{Number of configurations where each method achieves the highest Jaccard Index across spatial resolutions. Values represent winning counts from 40 possible noise/outlier combinations per resolution. 3C-FBI leads overall (44.4\%), dominating at moderate resolutions, while \citet{qi2024robust} (33.4\%) excels at high and resolutions.}
\label{tab:WinnerCount}
\makebox[1\textwidth][c]{\begin{tabular}{l|cccccccc|r}
\toprule
Method & $\infty$ & 480 & 240 & 160 & 80 & 40 & 20 & 12 & Total \\
\midrule
RHT - \citet{xu1990new} & 7 & 4 & 6 & 3 & 0 & 0 & 0 & 0 & 20 (6.3\%)\\
RCD - \citet{chen2001efficient} & 2 & 4 & 1 & 3 & 4 & 0 & 0 & 0 & 14 (4.4\%)\\
RFCA - \citet{ladron2011robust} & 1 & 0 & 0 & 0 & 1 & 4 & 2 & 1 & 9 (2.8\%)\\
\citet{nurunnabi2018robust} & 3 & 2 & 3 & 2 & 1 & 0 & 0 & 0 & 11 (3.4\%)\\
\citet{guo2019iterative} & 2 & 2 & 0 & 1 & 0 & 0 & 0 & 0 & 5 (1.6\%)\\
\citet{greco2023impartial} & 2 & 0 & 0 & 0 & 0 & 1 & 0 & 11 & 14 (4.4\%)\\
\textbf{\citet{qi2024robust}} & 16 & 17 & 15 & 12 & 8 & 2 & 14 & 22 & 106 (33.1\%)\\
\textbf{3C-FBI} & 7 & 11 & 15 & 19 & 26 & 33 & 24 & 6 & 141 (44.1\%) \\
\bottomrule
\end{tabular}}
\end{table*}

The performance exhibits a strong dependence on spatial resolution.
The discretization alters the ratio of inlier points and outlier points across different resolutions and contamination levels, revealing a notable asymmetric degradation pattern: the amount of inliers decreases much more rapidly with decreasing resolution compared to outliers (\ref{tab:detection_rates}).

\begin{table*}[htbp]
\footnotesize
\caption{Effect of image resolution on the mean number of inlier (In) and outlier (Out) points relative to the total points, across different resolutions and contamination levels.}
\label{tab:detection_rates}
\makebox[1\textwidth][c]{\begin{tabular}{l*{16}{r}}
\toprule
Outliers \% & \multicolumn{2}{c}{0} & \multicolumn{2}{c}{10} & \multicolumn{2}{c}{20} & \multicolumn{2}{c}{30} & \multicolumn{2}{c}{40} & \multicolumn{2}{c}{50} & \multicolumn{2}{c}{60} & \multicolumn{2}{c}{70} \\
Resolution & In & Out & In & Out & In & Out & In & Out & In & Out & In & Out & In & Out & In & Out \\
\midrule
$\infty$ & 100 & 0 & 90 & 10 & 80 & 20 & 70 & 30 & 60 & 40 & 50 & 50 & 40 & 60 & 30 & 70 \\
480×480 & 99.7 & 0 & 89.6 & 10.0 & 79.8 & 20.0 & 69.8 & 30.0 & 59.8 & 40.0 & 49.9 & 50.0 & 39.9 & 60.0 & 30.0 & 70.0 \\
240×240 & 98.5 & 0 & 89.0 & 10.0 & 79.0 & 20.0 & 69.4 & 30.0 & 59.5 & 40.0 & 49.7 & 50.0 & 39.7 & 60.0 & 29.9 & 69.9 \\
160×160 & 96.8 & 0 & 87.4 & 10.0 & 78.0 & 20.0 & 68.3 & 30.0 & 58.8 & 40.0 & 49.1 & 50.0 & 39.4 & 59.9 & 29.7 & 69.9 \\
80×80 & 88.1 & 0 & 80.8 & 10.0 & 72.7 & 19.9 & 64.2 & 29.9 & 55.7 & 39.9 & 47.1 & 49.7 & 38.0 & 59.6 & 28.9 & 69.5 \\
40×40 & 67.5 & 0 & 63.2 & 9.9 & 58.2 & 19.8 & 52.5 & 29.8 & 46.9 & 39.5 & 41.0 & 49.3 & 33.5 & 58.9 & 26.6 & 68.2 \\
20×20 & 38.3 & 0 & 37.1 & 9.9 & 35.2 & 19.6 & 34.1 & 28.8 & 31.5 & 37.9 & 28.7 & 47.2 & 25.1 & 55.6 & 21.5 & 64.3 \\
12×12 & 23.0 & 0 & 22.7 & 9.6 & 22.3 & 18.8 & 21.6 & 27.4 & 20.7 & 35.4 & 19.4 & 42.8 & 17.9 & 49.4 & 16.0 & 56.3 \\
\bottomrule
\end{tabular}}
\end{table*}

Notably, at 12×12 resolution with 30\% contamination, the number of inliers drops dramatically (from 70 to 21.6 points), while the number of outliers remains proportional to the contamination level (changing from 30 to 27.4 points). 
This resolution-dependent behavior suggests a critical threshold around 160×160 pixels, below which the number of inliers drop rapidly due to the circle edge covering so few pixels in the grid. 
Note that no correction for this drop in inliers is made in Figure \ref{fig:Roman2024_Jaccard_Heatmap_mean}.

\begin{figure*}[!htbp]
\begin{center}
    \centering
   \makebox[\textwidth][c]{\includegraphics[width=1.2\textwidth]{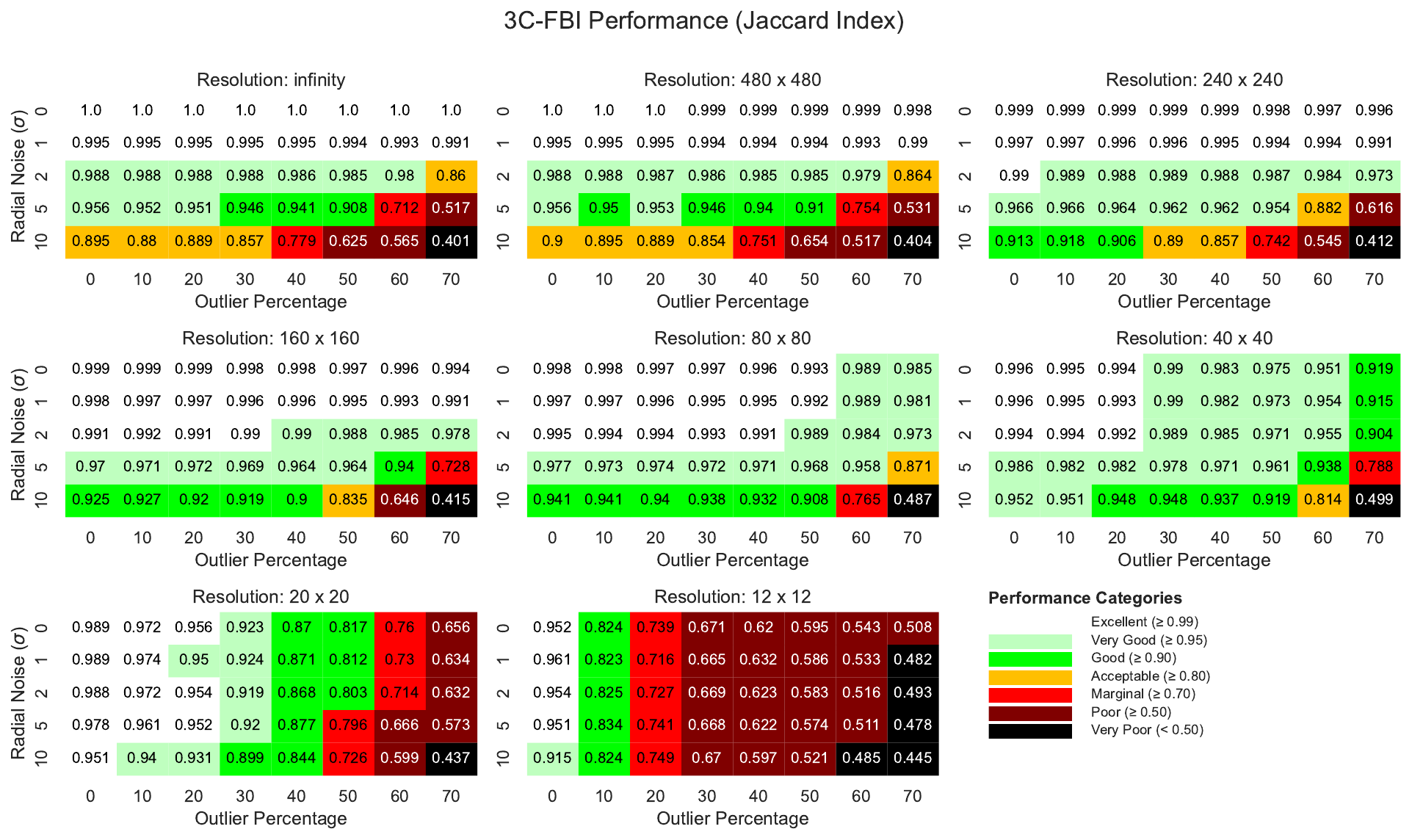}}
    \caption{3C-FBI performance measured by Jaccard Index across varying resolutions, noise levels, and outlier percentages. Color coding indicates performance categories from excellent (white/yellow, $\geq$0.95) to very poor (purple, $<$0.50). Performance remains very good at resolutions down to 80$\times$80 pixels with up to 60\% outliers, and radial noise up to 5\%,  but degrades significantly at lower resolutions.}
    \label{fig:Jaccard_3C-FBI}
\end{center}
\end{figure*}

Figure \ref{fig:Jaccard_3C-FBI} reveals robust algorithm performance across varying resolutions. Excellent results are maintained even at $40 \times 40$ resolution, with the majority of cases from infinite resolution down to $40 \times 40$ achieving at least Very Good performance (Jaccard Index $\geq 0.95$). At $20 \times 20$ resolution, while no excellent results are observed, approximately half the configurations maintain Good performance ($\geq 0.90$). The $12 \times 12$ resolution case demonstrates significant degradation, with only $12.5\%$ of configurations achieving Good performance and merely $25\%$ reaching Acceptable levels ($\geq 0.80$). Notably, across all resolutions, the combination of $10\%$ noise and $70\%$ outliers consistently produces Very Poor results ($< 0.50$), establishing this parameter combination as an effective performance boundary for the algorithm.

In summary, our evaluation demonstrates that 3C-FBI achieves state-of-the-art accuracy (Jaccard Index = 0.896) in real-world applications while maintaining high computational efficiency (40.31 fps). The method exhibits particular robustness under challenging conditions, including motion blur, variable lighting, and partial occlusions. It maintains stable performance up to 20\% outlier contamination and remains effective even under significant spatial discretization. Notably, 3C-FBI outperforms all other methods in 44.4\% of tested scenarios, dominating at moderate resolutions, while \citet{qi2024robust} leads at higher resolutions. These characteristics make 3C-FBI particularly well suited for real-time applications in robotics, medical imaging, and industrial quality control.

 \section{Discussion and Conclusion}
\label{sec:Conclusion}
This paper presented 3C-FBI, a circle detection algorithm combining combinatorial sampling with convolution-based feature extraction. 
Through comprehensive evaluation across real-world and synthetic scenarios, we demonstrated its effectiveness compared to both classical and state-of-the-art methods.

In real-world Parkinson's disease assessment data (Experiment A), 3C-FBI achieved a Jaccard Index of 0.896, matching the performance of recent state-of-the-art methods including \citet{qi2024robust} and \citet{greco2023impartial}, while significantly outperforming classical approaches like RHT (0.757) and RCD (0.620). 
When examining accuracy metrics in detail, our method showed comparable Average Distance (2.388 pixels) and RMSE (1.083) to both Qi et al. (2.353 pixels, 1.089) and Greco et al. (2.370 pixels, 1.082). 
The method demonstrated particular robustness under challenging imaging conditions, maintaining stable detection with motion blur and lighting variations. 
Most notably, 3C-FBI achieved this accuracy while maintaining real-time processing capability at 69.3 fps, which is slower than Greco et al.'s method (111.4 fps) and Qi et al.'s approach (106.9 fps), but substantially outperforming classical methods like RHT (26.8 fps) and RCD (39.5 fps).

In controlled synthetic testing with semicircular patterns (Experiment B1), 3C-FBI demonstrated exceptional robustness to outliers, maintaining a mean Jaccard Index of 0.989 across varying contamination levels. 
This performance closely matched modern approaches like Qi's method (0.991) and Ladrón de Guevara's RFCA (0.991), while significantly surpassing classical methods like RHT (0.964). 
The fact that all methods achieved a Jaccard index above 0.96 on average is in our opinion a sign that this test is too easy and lacks discriminatory power.

Our extended resolution analysis (Experiment B2) revealed clear patterns of method-specific strengths across varying spatial resolutions, noise levels, and outlier rates. 3C-FBI outperformed all other methods overall, achieving the highest Jaccard Index in 141 out of 320 configurations (44.1\%), and dominating at moderate to low resolutions, particularly from $240 \times 240$ down to $40 \times 40$ pixels. In contrast, \citet{qi2024robust} showed superior performance at the highest and lowest extremes, leading in 106 configurations (33.1\%) and excelling at both infinite and $12 \times 12$ pixel resolutions. Other methods, including RHT, RCD, and \citet{greco2023impartial}, showed more localized areas of competitiveness under specific resolution and degradation settings.

Importantly, performance heatmaps revealed that 3C-FBI maintains excellent to very good detection accuracy (Jaccard $\geq 0.95$) even with up to $60\%$ outliers and $5\%$ radial noise at resolutions down to $80 \times 80$ pixels. Below this threshold, performance degraded more significantly, particularly under high noise or extreme downsampling. These findings emphasize 3C-FBI’s robustness across practical imaging conditions, especially in moderate-resolution scenarios where real-time and reliable circle detection is most needed.

Future work should focus on enhancing low-resolution performance and high levels of outliers. 
The development of resolution-invariant  algorithms and extension to multi-circle detection scenarios represent promising directions for improvement. 
3C-FBI's combination of accuracy, efficiency, and robustness positions it as a valuable contribution to circle detection, particularly for applications requiring real-time processing under varying imaging conditions.

\section*{CRediT authorship contribution statement}

Esteban Roman Catafau: Methodology, Investigation, Writing – original draft, Validation.
Torbj\"orn Nordling: Conceptualization, Methodology, Writing – review \& editing, Supervision, Project administration.

\section*{Declaration of Competing Interest }
The authors declare that they have no known competing financial interests or personal relationships that could have appeared to influence the work reported in this paper.

\section*{Declaration of generative AI and AI-assisted technologies in the writing process}

During the preparation of this work the authors used ChatGPT in order to check English grammar. 
After using this tool/service, the authors reviewed and edited the content as needed and take full responsibility for the content of the publication.

\section*{Ethics and Consent}

This study involves human participants and has received ethical approval from the relevant ethics committee. 
\textbf{Ethical Approval}:
This research was conducted in accordance with the guidelines set forth by the Institutional Review Board of the National Cheng Kung University Hospital. 
Ethical approval was granted on 2019-12-20, with the approval number B-ER-108-362.
Our data was collected from 2020-02-01 to 2020-06-30.
\textbf{Participants Consent}:
Informed consent was obtained from all individual participants included in the study. 
Participants were fully informed about the purpose, procedures, risks, and benefits of the study, and consent was obtained before any data collection commenced.
\textbf{Data Confidentiality}:
All data collected from participants were kept confidential and used solely for the purposes of this study. Personal identifiers were removed to ensure the anonymity of the participants.

\section*{Acknowledgments}
\label{sec:acknowledgments}

This work was supported by the Ministry of Science and Technology of Taiwan through grants: MOST 108-2811-E-006-046, MOST 109-2224-F-006-003, MOST 110-2222-E-006-010, and MOST 111-2221-E-006-186 as well as National Science and Technology Council of Taiwan grant NSTC 112-2314–B-006-079 and NSTC 113-2314–B-006-069. 
The authors express their deep gratitude to Prof. Chi-Lun Lin, Prof. Chun-Hsiang Tan, Dr. Akram Ashyani, Yeh-Chin Ted, Jian-Hao Guo, Wei-Fang Tsai, Yu-Shan Lin, Hao-Wei Tu, Yu-Hsiang Su, and Chien-Chih Wang for developing the protocol and collected the data used in this study. 
The authors also thank Magdalena Valderrama for help with graphic design.

\bibliographystyle{unsrtnat}

\bibliography{References}

\begin{thebibliography}{31}
\providecommand{\natexlab}[1]{#1}
\providecommand{\url}[1]{\texttt{#1}}
\expandafter\ifx\csname urlstyle\endcsname\relax
  \providecommand{\doi}[1]{doi: #1}\else
  \providecommand{\doi}{doi: \begingroup \urlstyle{rm}\Url}\fi

\bibitem[Ma et~al.(2002)Ma, Wang, and Tan]{ma2002iris}
Li~Ma, Yunhong Wang, and Tieniu Tan.
\newblock Iris recognition using circular symmetric filters.
\newblock In \emph{2002 International Conference on Pattern Recognition},
  volume~2, pages 414--417. IEEE, 2002.

\bibitem[Van~Huan and Kim(2008)]{van2008novel}
Nguyen Van~Huan and Hakil Kim.
\newblock A novel circle detection method for iris segmentation.
\newblock In \emph{2008 Congress on Image and Signal Processing}, volume~3,
  pages 620--624. IEEE, 2008.

\bibitem[Arvacheh and Tizhoosh(2006)]{arvacheh2006iris}
Ehsan~Mohammadi Arvacheh and Hamid~R Tizhoosh.
\newblock Iris segmentation: Detecting pupil, limbus and eyelids.
\newblock In \emph{2006 International Conference on Image Processing}, pages
  2453--2456. IEEE, 2006.

\bibitem[Xavier et~al.(2005)Xavier, Pacheco, Castro, Ruano, and
  Nunes]{xavier2005fast}
Joao Xavier, Marco Pacheco, Daniel Castro, Ant{\' o}nio Ruano, and Urbano
  Nunes.
\newblock Fast line, arc/circle and leg detection from laser scan data in a
  player driver.
\newblock In \emph{Proceedings of the 2005 IEEE International Conference on
  Robotics and Automation}, pages 3930--3935. IEEE, 2005.

\bibitem[Glover and Bartolozzi(2016)]{glover2016event}
Arren Glover and Chiara Bartolozzi.
\newblock Event-driven ball detection and gaze fixation in clutter.
\newblock In \emph{2016 IEEE/RSJ International Conference on Intelligent Robots
  and Systems (IROS)}, pages 2203--2208. IEEE, 2016.

\bibitem[Teixid{\'o} et~al.(2012)Teixid{\'o}, Pallej{\`a}, Font, Tresanchez,
  Moreno, and Palac{\'\i}n]{teixido2012two}
Merc{\`e} Teixid{\'o}, Tom{\`a}s Pallej{\`a}, Davinia Font, Marcel Tresanchez,
  Javier Moreno, and Jordi Palac{\'\i}n.
\newblock Two-dimensional radial laser scanning for circular marker detection
  and external mobile robot tracking.
\newblock \emph{Sensors}, 12\penalty0 (12):\penalty0 16482--16497, 2012.

\bibitem[Koresh and Deva(2019)]{koresh2019computer}
MHJD Koresh and J~Deva.
\newblock Computer vision based traffic sign sensing for smart transport.
\newblock \emph{Journal of Innovative Image Processing (JIIP)}, 1\penalty0
  (01):\penalty0 11--19, 2019.

\bibitem[Satti et~al.(2020)Satti, Devi, Dhar, and
  Srinivasan]{satti2020enhancing}
Satish~Kumar Satti, K~Suganya Devi, Prasenjit Dhar, and P~Srinivasan.
\newblock Enhancing and classifying traffic signs using computer vision and
  deep convolutional neural network.
\newblock In \emph{Machine Learning, Image Processing, Network Security and
  Data Sciences: Second International Conference, MIND 2020, Silchar, India,
  July 30-31, 2020, Proceedings, Part I 2}, pages 243--253. Springer, 2020.

\bibitem[Ok and Ba{\c{s}}eski(2015)]{ok2015circular}
Ali~Ozgun Ok and Emre Ba{\c{s}}eski.
\newblock Circular oil tank detection from panchromatic satellite images: A new
  automated approach.
\newblock \emph{IEEE Geoscience and Remote Sensing Letters}, 12\penalty0
  (6):\penalty0 1347--1351, 2015.

\bibitem[Cai et~al.(2014)Cai, Sui, Lv, and Song]{cai2014automatic}
Xiaoyu Cai, Haigang Sui, Ruipeng Lv, and Zhina Song.
\newblock Automatic circular oil tank detection in high-resolution optical
  image based on visual saliency and hough transform.
\newblock In \emph{2014 IEEE Workshop on Electronics, Computer and
  Applications}, pages 408--411. IEEE, 2014.

\bibitem[Duda and Hart(1972)]{duda1972use}
Richard~O Duda and Peter~E Hart.
\newblock Use of the hough transformation to detect lines and curves in
  pictures.
\newblock \emph{Communications of the ACM}, 15\penalty0 (1):\penalty0 11--15,
  1972.

\bibitem[Xu et~al.(1990)Xu, Oja, and Kultanen]{xu1990new}
Lei Xu, Erkki Oja, and Pekka Kultanen.
\newblock A new curve detection method: randomized hough transform (rht).
\newblock \emph{Pattern recognition letters}, 11\penalty0 (5):\penalty0
  331--338, 1990.

\bibitem[Chen and Chung(2001)]{chen2001efficient}
Teh-Chuan Chen and Kuo-Liang Chung.
\newblock An efficient randomized algorithm for detecting circles.
\newblock \emph{Computer vision and image understanding}, 83\penalty0
  (2):\penalty0 172--191, 2001.

\bibitem[Atherton and Kerbyson(1999)]{atherton1999size}
Tim~J Atherton and Darren~J Kerbyson.
\newblock Size invariant circle detection.
\newblock \emph{Image and Vision computing}, 17\penalty0 (11):\penalty0
  795--803, 1999.

\bibitem[Ladr{\'o}n~de Guevara et~al.(2011)Ladr{\'o}n~de Guevara, Mu{\~n}oz,
  De~C{\'o}zar, and Bl{\'a}zquez]{ladron2011robust}
I~Ladr{\'o}n~de Guevara, Jos{\'e} Mu{\~n}oz, OD~De~C{\'o}zar, and Elidia~B
  Bl{\'a}zquez.
\newblock Robust fitting of circle arcs.
\newblock \emph{Journal of Mathematical Imaging and Vision}, 40:\penalty0
  147--161, 2011.

\bibitem[Nurunnabi et~al.(2018)Nurunnabi, Sadahiro, and
  Laefer]{nurunnabi2018robust}
Abdul Nurunnabi, Yukio Sadahiro, and Debra~F Laefer.
\newblock Robust statistical approaches for circle fitting in laser scanning
  three-dimensional point cloud data.
\newblock \emph{Pattern Recognition}, 81:\penalty0 417--431, 2018.

\bibitem[Guo and Yang(2019)]{guo2019iterative}
Jianfeng Guo and Jiameng Yang.
\newblock An iterative procedure for robust circle fitting.
\newblock \emph{Communications in Statistics-Simulation and Computation},
  48\penalty0 (6):\penalty0 1872--1879, 2019.

\bibitem[Greco et~al.(2023)Greco, Pacillo, and Maresca]{greco2023impartial}
Luca Greco, Simona Pacillo, and Piera Maresca.
\newblock An impartial trimming algorithm for robust circle fitting.
\newblock \emph{Computational Statistics \& Data Analysis}, 181:\penalty0
  107686, 2023.

\bibitem[Qi et~al.(2024)Qi, Wang, Luo, Cheng, and Liu]{qi2024robust}
Zhijun Qi, Wei Wang, Tao Luo, Wenjie Cheng, and Zengquan Liu.
\newblock A robust circle fitting method for component fiducialization.
\newblock \emph{Nuclear Instruments and Methods in Physics Research Section A:
  Accelerators, Spectrometers, Detectors and Associated Equipment},
  1068:\penalty0 169775, 2024.

\bibitem[Al-Sharadqah and Chernov(2009)]{al2009error}
Ali Al-Sharadqah and Nikolai Chernov.
\newblock Error analysis for circle fitting algorithms.
\newblock 2009.

\bibitem[Wang and Suter(2003)]{wang2003using}
Hanzi Wang and David Suter.
\newblock Using symmetry in robust model fitting.
\newblock \emph{Pattern Recognition Letters}, 24\penalty0 (16):\penalty0
  2953--2966, 2003.

\bibitem[Cai et~al.(2013)Cai, Man, Wang, Chen, and Yuan]{cai2013combined}
Guo~Zhu Cai, Kai~Di Man, Shao~Ming Wang, Wen~Jun Chen, and Jian~Dong Yuan.
\newblock A combined application of laser tracker and spatialanalyzer in
  alignment of accelerator.
\newblock \emph{Applied Mechanics and Materials}, 333:\penalty0 58--61, 2013.

\bibitem[Ayala-Ramirez et~al.(2006)Ayala-Ramirez, Garcia-Capulin, Perez-Garcia,
  and Sanchez-Yanez]{ayala2006circle}
Victor Ayala-Ramirez, Carlos~H Garcia-Capulin, Arturo Perez-Garcia, and Raul~E
  Sanchez-Yanez.
\newblock Circle detection on images using genetic algorithms.
\newblock \emph{Pattern Recognition Letters}, 27\penalty0 (6):\penalty0
  652--657, 2006.

\bibitem[Jia et~al.(2011)Jia, Peng, Liu, and Wang]{jia2011fast}
Li-qin Jia, Cheng-zhang Peng, Hong-min Liu, and Zhi-heng Wang.
\newblock A fast randomized circle detection algorithm.
\newblock In \emph{2011 4th International Congress on Image and Signal
  Processing}, volume~2, pages 820--823. IEEE, 2011.

\bibitem[Chung et~al.(2012)Chung, Huang, Shen, Krylov, Yurin, and
  Semeikina]{chung2012efficient}
Kuo-Liang Chung, Yong-Huai Huang, Shi-Ming Shen, Andrey~S Krylov, Dmitry~V
  Yurin, and Ekaterina~V Semeikina.
\newblock Efficient sampling strategy and refinement strategy for randomized
  circle detection.
\newblock \emph{Pattern recognition}, 45\penalty0 (1):\penalty0 252--263, 2012.

\bibitem[Lestriandoko and Sadikin(2016)]{Lestriandoko2017Circle}
Nova~Hadi Lestriandoko and Rifki Sadikin.
\newblock Circle detection based on hough transform and mexican hat filter.
\newblock In \emph{2016 International conference on computer, control,
  informatics and its applications (IC3INA)}, pages 153--157. IEEE, 2016.

\bibitem[Ferede et~al.(2019)Ferede, Xie, Jin, Du, and Shi]{ferede2019channel}
Shambel Ferede, Xuemei Xie, Xing Jin, Jiang Du, and Guangming Shi.
\newblock Channel feature enhanced detector for small ball detection.
\newblock In \emph{Chinese Conference on Pattern Recognition and Computer
  Vision (PRCV)}, pages 3--13. Springer, 2019.

\bibitem[Kamble et~al.(2019)Kamble, Keskar, and
  Bhurchandi]{kamble2019convolutional}
Paresh~R Kamble, Avinash~G Keskar, and Kishor~M Bhurchandi.
\newblock A convolutional neural network based 3d ball tracking by detection in
  soccer videos.
\newblock In \emph{Eleventh International Conference on machine vision (ICMV
  2018)}, volume 11041, page 110412O. International Society for Optics and
  Photonics, 2019.

\bibitem[Rom{\'a}n~Catafau and Nordling()]{romancibica}
Esteban Rom{\'a}n~Catafau and Torbj{\"o}rn Nordling.
\newblock Cibica: Circle identification in blurry images using combinatorial.
\newblock \emph{Available at SSRN 4542991}.

\bibitem[Goetz et~al.(2008)Goetz, Tilley, Shaftman, Stebbins, Fahn,
  Martinez-Martin, Poewe, Sampaio, Stern, Dodel, et~al.]{goetz2008movement}
Christopher~G Goetz, Barbara~C Tilley, Stephanie~R Shaftman, Glenn~T Stebbins,
  Stanley Fahn, Pablo Martinez-Martin, Werner Poewe, Cristina Sampaio,
  Matthew~B Stern, Richard Dodel, et~al.
\newblock Movement disorder society-sponsored revision of the unified
  parkinson's disease rating scale (mds-updrs): scale presentation and
  clinimetric testing results.
\newblock \emph{Movement disorders: official journal of the Movement Disorder
  Society}, 23\penalty0 (15):\penalty0 2129--2170, 2008.

\bibitem[Ashyani et~al.(2022)Ashyani, Lin, Roman, Yeh, Kuo, Tsai, Lin, Tu, Su,
  Wang, Tan, and Nordling]{ashyani2022digitization}
Akram Ashyani, Chi-Lun Lin, Esteban Roman, Ted Yeh, Tachyon Kuo, Wei-Fang Tsai,
  Yushan Lin, Ric Tu, Austin Su, Chien-Chih Wang, Chun-Hsiang Tan, and Torbj{
  \"o}rn E~M Nordling.
\newblock Digitization of updrs upper limb motor examinations towards automated
  quantification of symptoms of parkinson's disease.
\newblock \emph{Manuscript in preparation}, 2022.

\end{thebibliography}

\appendix
\renewcommand{\thefigure}{\arabic{figure}} \setcounter{figure}{14} \section{Appendix}
\subsection{Additional Results for the experiment B2}  
\begin{figure*}[!t]
\begin{center}
	\centering
	\makebox[\textwidth][c]{\includegraphics[width=1.2\textwidth]{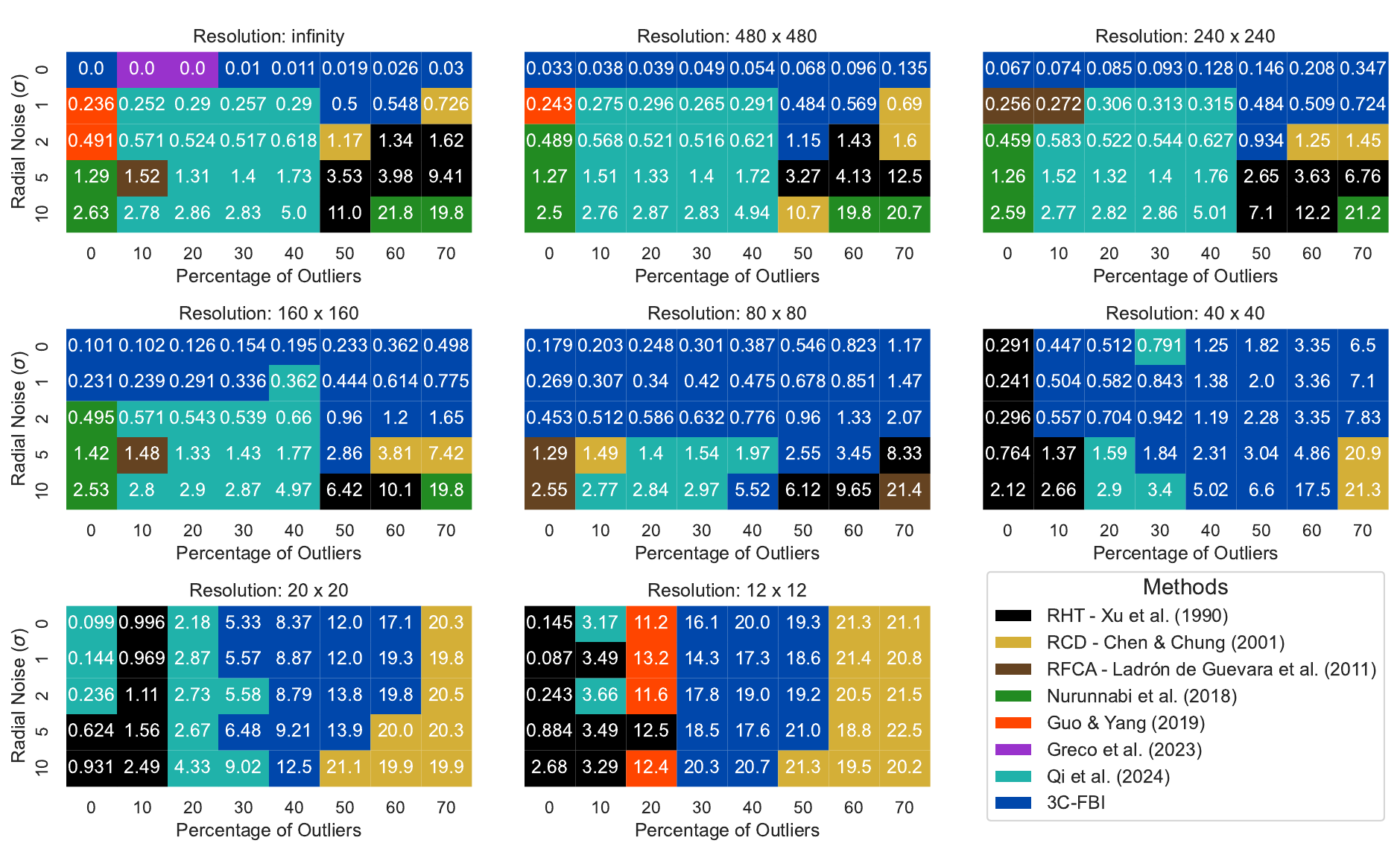}}
	\caption{Average Distance (AD) performance comparison across partition resolutions, radial noise levels, and outlier percentages. Values represent mean distance in millimeters between estimated and true circle parameters For 100 realizations.}
	\label{fig:Roman2024_AD_Heatmap_mean}
\end{center}
\end{figure*}

Average Distance measurements (Figure~\ref{fig:Roman2024_AD_Heatmap_mean}) reveal increasing error with noise and outliers at all  partition resolutions. 
At infinite resolution, errors remain below 1mm until reaching 50\% outliers or 5\% noise, showing the method's stability under moderate conditions.

\begin{figure*}[!p]
\begin{center}
	\centering
	\makebox[\textwidth][c]{\includegraphics[width=1.2\textwidth]{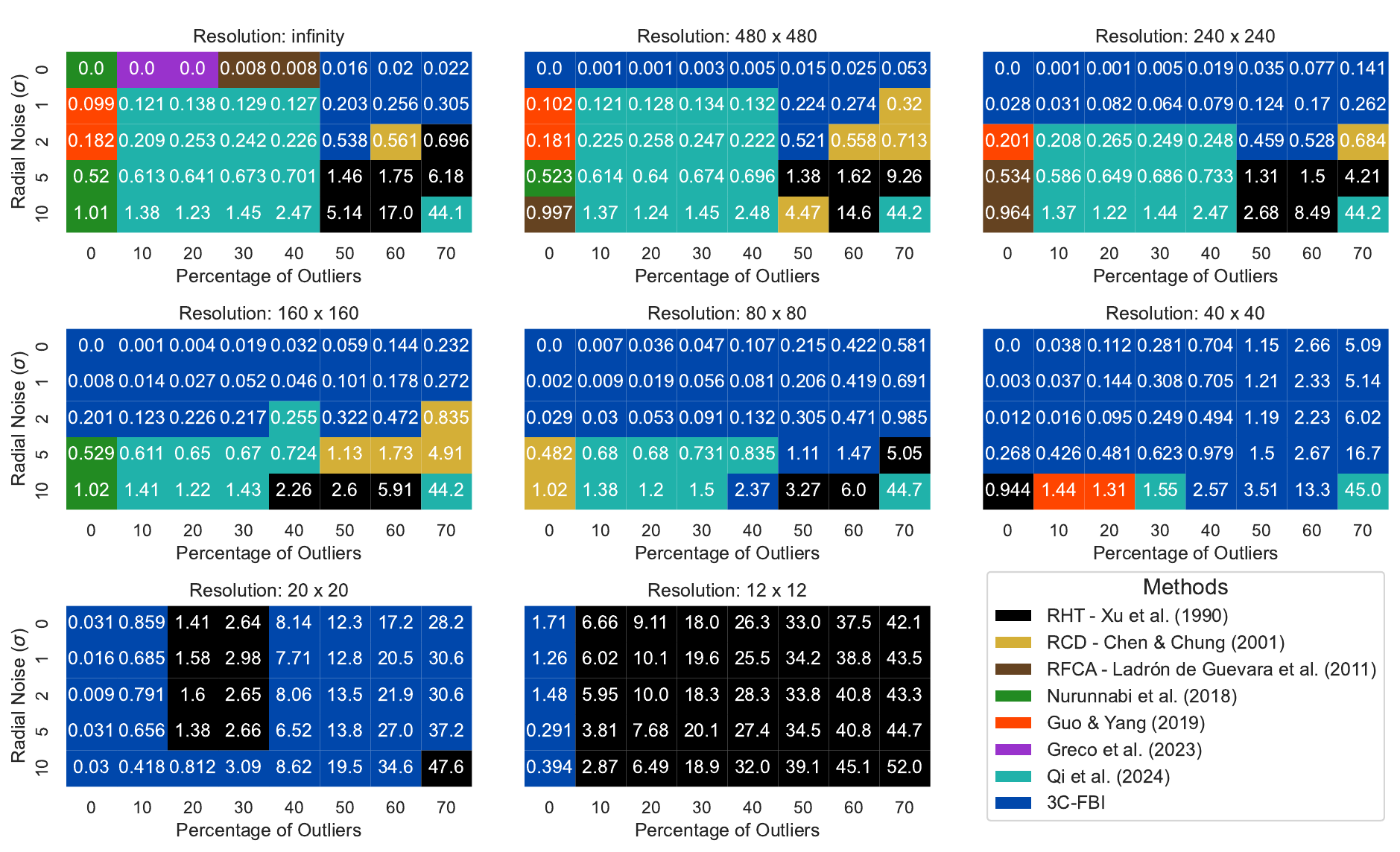}}
	\caption{Root Mean Square Error (RMSE) performance comparison across partition resolutions, noise levels, and outlier percentages. Values show mean error in millimeters between detected and true circle parameters. Noise ratio is defined as Radial
standard deviation/Radius of circle.}
	\label{fig:Roman2024_RMSE_Heatmap_mean}
\end{center}
\end{figure*}

RMSE analysis (Figure~\ref{fig:Roman2024_RMSE_Heatmap_mean}) shows similar patterns but highlights the exponential increase in error at extreme conditions. Performance remains stable (RMSE $<0.5$ mm) at low partition resolutions with up to 20\% outliers, but deteriorates rapidly at higher resolutions and noise levels.

\subsection{Analytic Derivation of Circle Through Three Points}

Given three non-collinear points \( P_1 = (x_1, y_1) \), \( P_2 = (x_2, y_2) \), and \( P_3 = (x_3, y_3) \), the unique circle passing through them has center \((x_c, y_c)\) and radius \(r\) defined as:

\begin{equation}
\begin{aligned}
D &= 2 \left( x_1(y_2 - y_3) + x_2(y_3 - y_1) + x_3(y_1 - y_2) \right) \\
x_c &= \frac{(x_1^2 + y_1^2)(y_2 - y_3) + (x_2^2 + y_2^2)(y_3 - y_1) + (x_3^2 + y_3^2)(y_1 - y_2)}{D} \\
y_c &= \frac{(x_1^2 + y_1^2)(x_3 - x_2) + (x_2^2 + y_2^2)(x_1 - x_3) + (x_3^2 + y_3^2)(x_2 - x_1)}{D} \\
r &= \sqrt{(x_1 - x_c)^2 + (y_1 - y_c)^2}
\end{aligned}
\tag{A1}\label{eq:circle_formula}
\end{equation}

\end{document}